\documentclass{article}

\PassOptionsToPackage{numbers, compress}{natbib}

\usepackage[preprint]{neurips_2025}

\usepackage[utf8]{inputenc} 
\usepackage[T1]{fontenc}    
\usepackage{hyperref}       
\usepackage{url}            
\usepackage{booktabs}       
\usepackage{amsfonts}       
\usepackage{nicefrac}       
\usepackage{microtype}      
\usepackage{graphicx}       
\usepackage{amsmath}
\usepackage[dvipsnames]{xcolor}

\usepackage{graphicx}
\usepackage{amsmath}
\usepackage{amssymb}
\usepackage{booktabs}

\usepackage{multirow}
\usepackage{algorithm}
\usepackage[T1]{fontenc}
\usepackage{multicol}
\usepackage{stfloats}
\usepackage{xcolor}

\usepackage{mathtools}

\usepackage{makecell}
\usepackage{scalerel}
\usepackage{array}
\usepackage{dsfont}
\usepackage{url}
\usepackage{pifont}
\usepackage{stmaryrd}
\usepackage{adjustbox}
\usepackage{hyperref}
\usepackage{cleveref}

\usepackage[shortlabels]{enumitem}

\DeclareMathOperator*{\argmin}{arg\,min}
\newcolumntype{C}[1]{>{\centering\let\newline\\\arraybackslash\hspace{0pt}}p{#1}}
\newcolumntype{R}[1]{>{\raggedleft\let\newline\\\arraybackslash\hspace{0pt}}p{#1}}
\newcolumntype{L}[1]{>{\raggedright\let\newline\\\arraybackslash\hspace{0pt}}p{#1}}
\newcolumntype{M}[1]{>{\centering\let\newline\\\arraybackslash\hspace{0pt}}m{#1}}

\newcommand\mydots{\hbox to 1em{.\hss.\hss.}}

\definecolor{myblue}{RGB}{0, 250, 0}
\definecolor{mypink}{RGB}{237, 2, 140}
\definecolor{green1}{RGB}{148, 193, 117}
\definecolor{green2}{RGB}{90, 138, 57}
\definecolor{red}{RGB}{200, 20, 0}
\definecolor{blue}{RGB}{0, 20, 200}
\definecolor{grey}{RGB}{100, 100, 100}

\newcolumntype{D}[2]{%
    >{\adjustbox{angle=#1,lap=\width-(#2)}\bgroup}%
    l%
    <{\egroup}%
}

\usepackage{algpseudocode}  

\usepackage{scalerel,xparse}
\NewDocumentCommand\emojifire{}{\scalerel*{\includegraphics{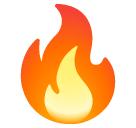}}{X}}
\NewDocumentCommand\emojifreeze{}{\scalerel*{\includegraphics{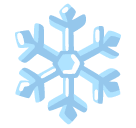}}{X}}

\title{Pts3D-LLM: Studying the Impact of Token Structure for 3D Scene
Understanding With Large Language Models}

\author{%
  Hugues Thomas \\
  Apple \\
  \texttt{hthomas24@apple.com} \\
  \And
  Chen Chen \\
  Apple \\
  \texttt{cchen64@apple.com} \\
  \And
  Jian Zhang \\
  Apple \\
  \texttt{jianz@apple.com} \\
}

\begin{document}

\maketitle

%
%

\begin{figure}[h]
    \vspace{-4ex}
    \centering
    \adjincludegraphics[width=0.99\textwidth,trim={{.0\width} {.361\height} {.015\width} {.0\height}},clip]{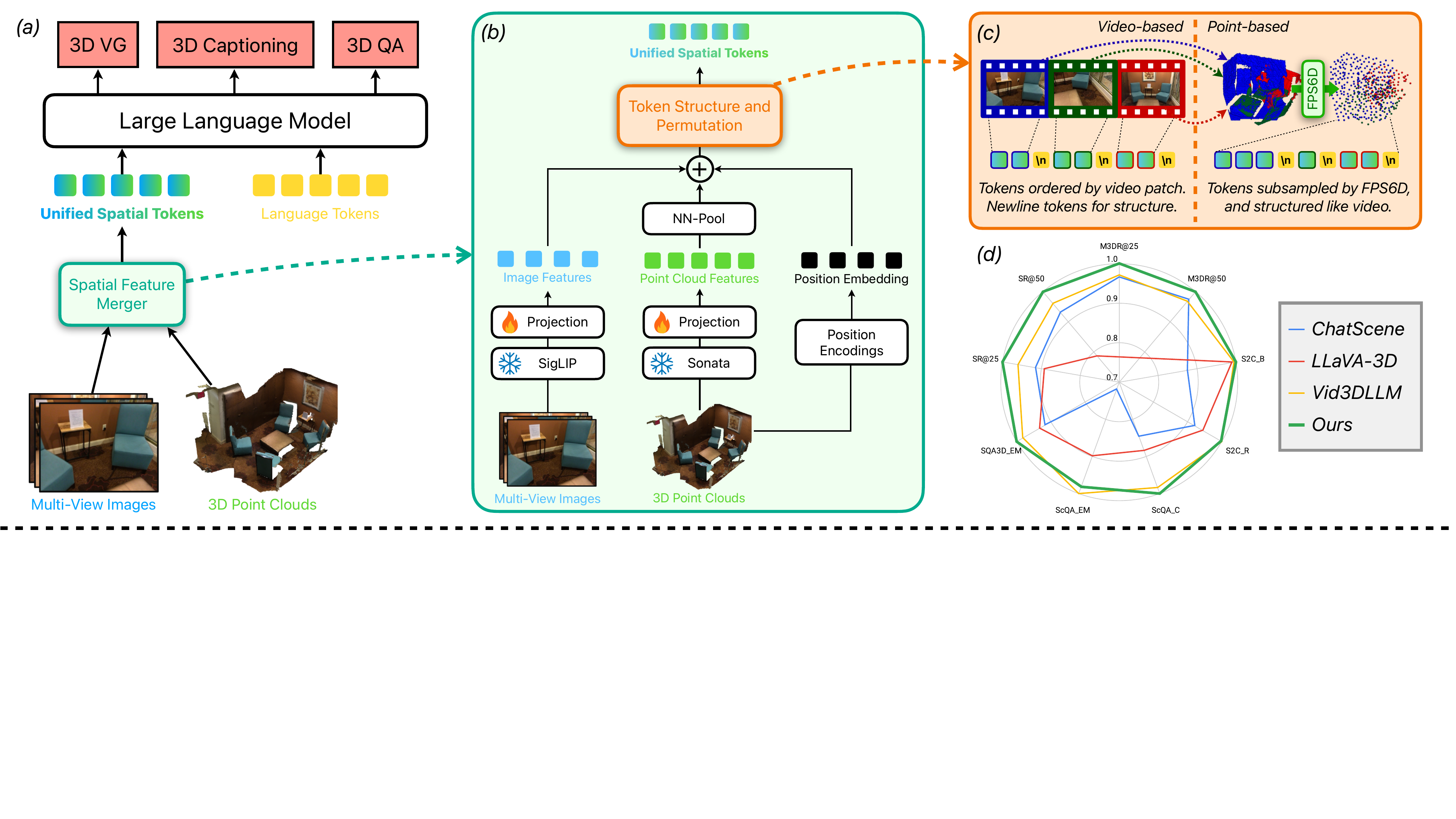}
    \caption{\textbf{Overview of our approach.} (a) We solve 3D scene understanding tasks with a multimodal LLM that merges image semantics, shape patterns, and location information in unified spatial tokens. (b) These tokens are built from SigLIP \citep{zhai2023sigmoid} (image encoder) features, Sonata \citep{sonata} (point cloud encoder) features, and position encodings. The point cloud features are pooled to the image feature locations (obtained from corresponding depths) via nearest neighbor interpolation. (c) The tokens' structure and permutation directly impact performance. In addition to the standard video-based structure, we propose a view-sensitive subsampling method (FPS6D) to obtain efficient point-based token structures. (d) Our approach achieves state-of-the-art performance across various 3D scene understanding benchmarks. Our code will be available at: {\color{mypink}\url{https://github.com/apple/ml-pts3dllm}}.}
    \label{fig:intro}
\end{figure}

\begin{abstract}

Effectively representing 3D scenes for Multimodal Large Language Models (MLLMs) is crucial yet challenging. Existing approaches commonly only rely on 2D image features and use varied tokenization approaches. This work presents a rigorous study of 3D token structures, systematically comparing video-based and point-based representations while maintaining consistent model backbones and parameters. We propose a novel approach that enriches visual tokens by incorporating 3D point cloud features from a Sonata pretrained Point Transformer V3 encoder. Our experiments demonstrate that merging explicit 3D features significantly boosts performance. Furthermore, we show that point-based token structures can rival video-based ones when the points are cleverly sampled and ordered. Our best models from both structures achieve state-of-the-art results on multiple 3D understanding benchmarks. We emphasize our analysis of token structures as a key contribution, alongside transparent reporting of results averaged over multiple seeds, a practice we believe is vital for robust progress in the field.

\end{abstract}

%
%

\section{Introduction}

Multimodal Large Language Models (MLLMs) have transformed how we approach complex tasks by integrating diverse data types, such as images and audio, alongside language. While the computer vision field has seen remarkable progress in 2D visual understanding, extending these capabilities to 3D scene understanding presents unique challenges, primarily due to the complexity of spatial relationships and the scarcity of large-scale 3D datasets. Early attempts at 3D MLLMs have shown promise in tasks like 3D Question Answering (3DQA), 3D Captioning (3DCap), and 3D Visual Grounding (3DVG). Two key stages are typically involved in the success of 3D MLLMs: encoding rich 3D scene features and structuring these features into a sequence of tokens that the transformer backbone can process alongside textual input.

Despite recent advancements, we identify two main limitations hindering further progress. Firstly, many contemporary methods, including top-performing ones \citep{leo, video3dllm}, predominantly rely on 2D image features. This is understandable given the larger size of 2D image datasets and the availability of image sequences with 3D benchmarks. However, this approach inherently overlooks explicit 3D structural information crucial for nuanced understanding. Our first key contribution directly addresses this by proposing a novel method to enrich visual tokens through the fusion of 2D image features with explicit 3D geometric features. These 3D features are extracted using the powerful, pretrained Sonata encoder \citep{sonata}, built upon the Point Transformer V3 (PTv3) architecture \citep{wu2024point}, leading to a more comprehensive scene representation.

Secondly, a significant and often overlooked issue is that there's no clear agreement or in-depth understanding of the best ways to structure these tokens. Current methods employ different strategies: LLaVA-3D \citep{llava3d} projects image features onto 3D voxels; Video-3D LLM \citep{video3dllm} treats image sequences as video, augmenting them with 3D position encodings; while others like Chat-Scene \citep{chat_scene} or LEO \citep{leo} use object-centric tokens derived from prior scene segmentation. Comparing these token structures is challenging because current approaches rely on varying model backbones and experimental conditions. Our second major contribution is a methodical comparison of different 3D token structures, specifically video-based versus point-based, using a consistent experimental setup. By keeping the model backbone and parameters consistent, we isolate and analyze the direct impact of token structure on performance. We intentionally exclude object-centric approaches to focus on scene-level feature encoding without reliance on intermediate segmentation steps.

Within our investigation of point-based structures, we identified that standard subsampling methods (like 3D Farthest Point Sampling or voxel averaging) often compromise performance by either discarding crucial viewpoint information or by averaging features from different views, potentially blurring distinctive details. To overcome this, our third key contribution is the introduction of FPS6D, a novel view-sensitive point sampling strategy. FPS6D operates in a 6-dimensional space, considering both the 3D coordinates of a point and the 3D coordinates of the camera view origin from which it is observed. This ensures that the selected tokens are not only spatially representative but also capture information from a diverse set of viewpoints, leading to a more informative and efficient token sequence for the LLM. We demonstrate that point-based structures, when intelligently sampled and ordered using FPS6D, can rival and even surpass the performance of established video-based structures, while offering greater flexibility.

We build upon the robust and open-source Video-3D LLM \citep{video3dllm} framework, extending its architecture to incorporate the PTv3-Sonata encoder and enabling flexible projection between point cloud and video feature structures. Our comprehensive experiments validate our approach: the integration of explicit 3D features provides a significant performance boost. Furthermore, our FPS6D-enhanced point-based models demonstrate remarkable competitiveness. Crucially, our best models, leveraging these innovations, achieve new state-of-the-art results on multiple challenging 3D scene understanding benchmarks. We also underscore the importance of transparent and robust evaluation by reporting results averaged over multiple random seeds, in addition to best model performance.

Our contributions are thus:
\begin{enumerate}
\item A novel method for MLLMs that enriches visual tokens by fusing 2D image semantics with explicit 3D geometric features from the Sonata point cloud encoder, demonstrating significant performance gains.
\item A systematic and fair comparison of 3D token structures (video-based vs. point-based), isolating their impact on performance by maintaining consistent model backbones and training parameters.
\item The introduction of FPS6D, a novel view-sensitive point sampling strategy for point-based tokenization, which balances spatial coverage and viewpoint diversity to create highly informative and efficient token sequences.
\item State-of-the-art performance on multiple 3D understanding benchmarks, coupled with transparent and reproducible results through multi-seed averaging.
\end{enumerate}

%
%

\section{Related Work}

\textbf{From 2D to 3D Large Multimodal Models}. The field of Large Multimodal Models (LMMs) has rapidly evolved, initially focusing on integrating 2D images with Large Language Models (LLMs). Seminal works like LLaVA \citep{llava} and BLIP-2 \citep{blip2} demonstrated effective methods for aligning visual features from a single image with language, using techniques ranging from simple projection layers to sophisticated Q-Former architectures. Recognizing the limitations of single-image inputs for real-world tasks, research extended towards multi-image understanding. This includes Video LMMs \citep{video_llama,li2023videochat,llama_adapterv2,video_chatgpt} processing frame sequences and early attempts at using multi-view images for 3D spatial reasoning \citep{3dllm,scenellm}. However, many multi-view approaches relied on implicit 3D learning within 2D LMM frameworks \citep{videollava}. More recent works like LLaVA-3D \citep{llava3d} and 3D-LLM \citep{3dllm} began explicitly modeling the 3D world from multi-view inputs, paving the way for deeper 3D spatial understanding.

\textbf{3D Scene Representation Strategies for LLMs}. A key challenge in 3D LMMs is how to effectively represent the 3D scene for the LLM. Various strategies have emerged. Some methods directly process scene-level point clouds using specialized encoders \citep{ll3da} or leverage Q-Former architectures \citep{ll3da,gpt4point} to bridge modalities. Others adopt an object-centric approach, first segmenting or detecting objects in the 3D point cloud \citep{leo,chat_scene} or from multi-view images \citep{3dllm,scenellm} and then feeding object-level features (either 3D point features \citep{leo,chat_scene} or aggregated 2D features \citep{3dllm,scenellm}) into the LLM. LLaVA-3D \citep{llava3d} proposed constructing "3D Patches" by aggregating 2D patch features within a 3D spatial context using positional embeddings. Scene-LLM \citep{scenellm} lifts multi-view 2D features into a 3D representation. These diverse approaches result in different token structures (e.g., object-based, voxel-based, projected 2D patches). However, comparing the efficacy of these structures is difficult due to variations in model backbones, training data, and parameters across studies. Our work directly addresses this by systematically categorizing and comparing different tokenization structures under a controlled experimental setup.

\textbf{Leveraging 2D Foundations and Combining Modalities}. Many 3D LMMs build upon pre-trained 2D vision-language models \citep{blip2,llava3d} or leverage powerful 2D foundation models to extract features from multi-view images \citep{leo,3dllm}. Techniques like incorporating 3D positional embeddings aim to imbue 2D features with spatial awareness. For instance, ODIN \citep{odin} used distinct positional encodings for 2D and 3D features to enable joint training. LLaVA-3D \citep{llava3d} similarly integrated 3D position-aware features into a 2D LMM framework. While leveraging strong 2D priors is beneficial, relying solely on image features can limit the capture of fine-grained 3D geometry. Our approach explores combining complementary features by integrating both image-based features and explicit geometric features derived directly from point clouds using a modern point cloud encoder (Point Transformer V3 \citep{wu2024point} with Sonata pertaining \citep{sonata}).

\textbf{Video-based Representations for 3D Understanding}. An alternative perspective treats multi-view images as a sequence, leveraging advancements in Video LLMs \citep{llava_video}. Works like LLaVA-OneVision \citep{llava_onevision} and Oryx MLLM \citep{oryx} have adapted video models for 3D question answering by fine-tuning on relevant datasets. However, these often lack detailed 3D spatial information integration. Recent work explicitly proposes enhancing video representations with 3D coordinate information to better utilize pre-trained 2D Video LLMs while capturing spatial structure \citep{video3dllm}. Our analysis investigates the effectiveness of such video-based token structures, comparing them against other 3D representation methods and finding them surprisingly effective despite potential token redundancy.

%
%

\section{Methodology}

\subsection{Overview}

We propose a 3D understanding architecture derived from the Video-3D LLM \citep{video3dllm} approach. Unless stated otherwise in this paper, we use the same methods and parameters; in particular, the LLM model, the vision encoder, the datasets, and the evaluation setups are kept identical. Our method focuses on the two core steps enabling 3D understanding: feature encoding and token structuring. For encoding, we combine image features (from the pretrained image model) with 3D point cloud features extracted using the Sonata encoder \citep{sonata}. We also fuse position encodings to the token as defined in \citep{video3dllm}. For token structuring, we explore multiple variants of the video-based and point-based structures to see which performs best. We keep the transformer backbone and parameters consistent across experiments to ensure fair comparisons.

Following Video-3D LLM, we use a SigLIP Vision Transformer \citep{zhai2023sigmoid} to obtain visual embeddings at each frame. Formally, we obtain patch-wise features $\mathcal{F_\mathrm{im}} \in \mathbb{R}^{V \times H \times W \times D}$ for which we can have corresponding 3D coordinates $\mathcal{C} \in \mathbb{R}^{V \times H \times W \times 3}$. $V$ is the number of camera views, and $H \times W$ is the number of vision transformer patches. We also use the same sinusoidal position encodings as in Video-3D LLM, $\mathcal{F_\mathrm{pe}} \in \mathbb{R}^{V \times H \times W \times D}$. We refer the reader to \citep{video3dllm} for more details. This arrangement of $V \times H \times W$ tokens defines our first video-based structure.

\begin{figure}[b]
    \centering
    \adjincludegraphics[width=0.99\textwidth,trim={{.0\width} {.536\height} {.015\width} {.0\height}},clip]{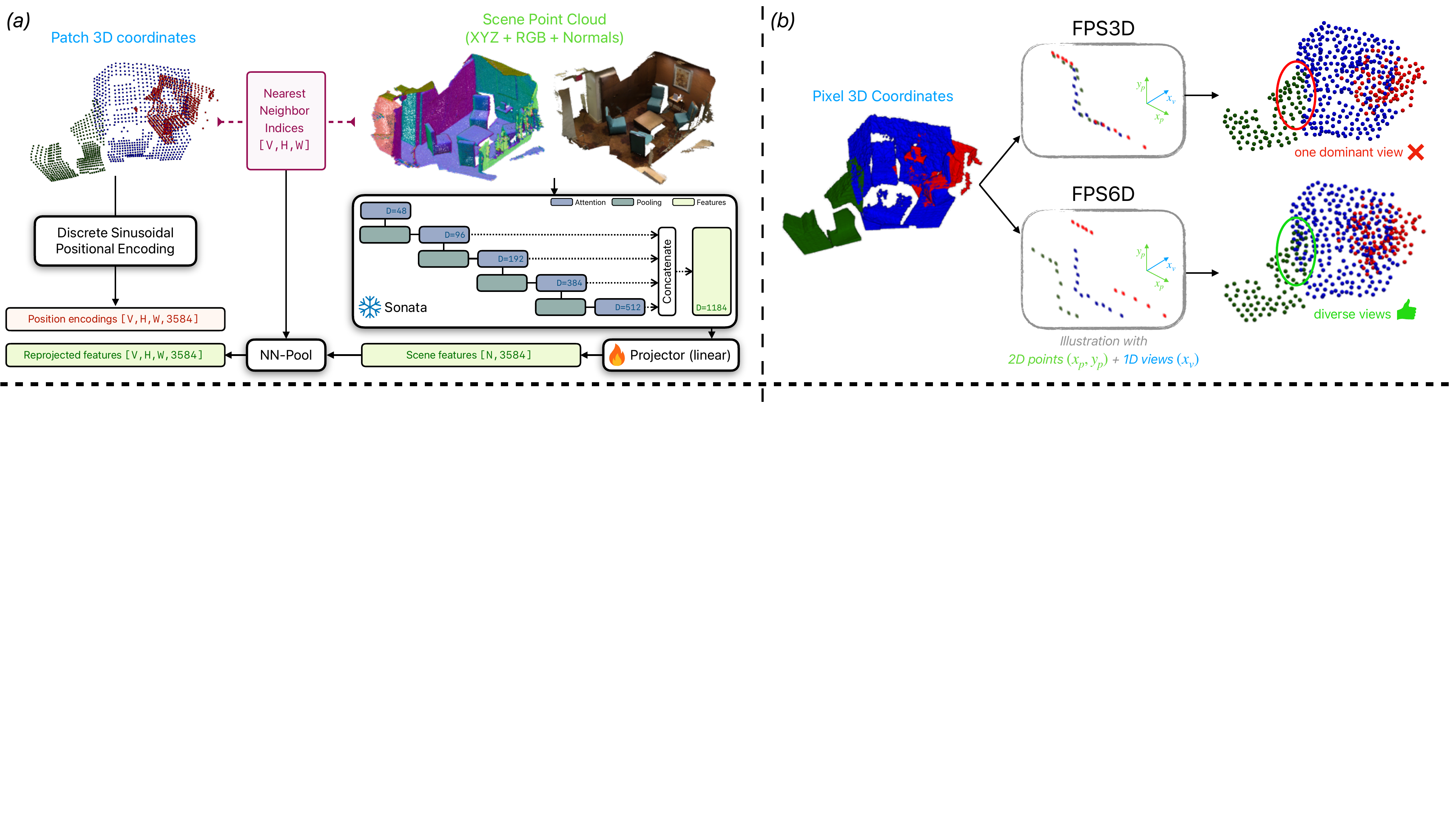}
    \caption{(a) \textbf{Illustration of our 3D encoder integration}. The scene point cloud is processed with Sonata to produce features by concatenating the four last layer outputs. The features are then projected to the token dimension with a linear layer. These scene features are pooled to the same location as the video patch features using a nearest neighbor pooling. (b) \textbf{Illustration of our FPS6D subsampling}. The pixel 3D coordinates are elevated in a 6-dimensional space by appending the camera view positions. As shown in a 2D + 1D space for clarity, FPS6D will maximize the diversity of views by selecting points regularly in this higher-dimensional space. On the real point cloud, we notice areas where FPS6D, as opposed to FPS3D, selects points from diverse views.}
    \label{fig:method}
\end{figure}

\subsection{3D Encoder Integration}

Our first contribution is to integrate 3D encoder features into the vision tokens fed to the LLM backbone. These new unified spatial tokens thus combine: (1) 2D image encoder features, which provide semantic information; (2) 3D point cloud encoder features, which provide additional semantic and shape pattern information; (3) position encodings, which describe the location of each token. As illustrated in \Cref{fig:method} (a), we use a pretrained Sonata encoder, which is pretrained on large-scale point cloud data and built on an encoder-only PTv3 architecture. This 5-layer encoder takes a point cloud with colors and normals as input and returns embeddings at each layer. The input point cloud is subsampled at each layer to reduce the number of points and increase the feature dimensions. We can upsample the features from the last 4 layers and concatenate them to obtain rich features at each point of the scene. These features can then be projected to the token dimension with a linear layer.

We obtain a set of features $\mathcal{F}_\mathrm{3d} \in \mathbb{R}^{N \times D}$ located at the coordinates of the scene point cloud $\mathcal{P} \in \mathbb{R}^{N \times 3}$, where $N$ is the number of points. To merge them with the vision embeddings, we therefore need to project these features to their corresponding patch. We use a nearest neighbor projection defined as:

\begin{equation}
\forall p \in \mathbb{R}^3, \quad \mathrm{NN}_\mathcal{P}(p) = \argmin_{i \in \{0, \dots, N-1\}} \left\| \mathcal{P}(i) - p \right\|    \:.
\end{equation}

Therefore, for a patch $(u, v)$ in the $k$-th view, the corresponding 3D scene point index is $\mathrm{NN}_\mathcal{P}(c(k, u, v)) \in \{0, \dots, N-1\}$; and our final feature values $\mathcal{F_\mathrm{token}} \in \mathbb{R}^{V \times H \times W \times D}$ are computed as: 

\begin{equation}
\label{eq:fvideo}
f_{token}\left(k, u, v\right) = f_{im}\left(k, u, v\right) + f_{3d}\left(\mathrm{NN}_\mathcal{P}(c(k, u, v))\right) + f_{pe}\left(k, u, v\right)   \:,
\end{equation}

\subsection{3D Token Structure}

The token embeddings as defined in \cref{eq:fvideo} follow a video-based structure. This structure is quite rigid and does not offer much flexibility apart from the choice of camera views. We refer to \citep{video3dllm} for a study of smarter selections of views. In the following, unless stated otherwise, we use a uniform selection of 32 views, which was their best setup. We believe that overlapping views are one of the reasons why video-based structure performs well, as it allows the LLM to get descriptions of the same location from multiple views, and to pay attention to the best ones.

Point-based structures are usually derived from video-based structures as subsampled versions of the full sequence of video tokens. Standard 3D subsampling methods do not take into account views and only focus on selecting points regularly in space. Voxel-average subsampling, as proposed by LLaVA-3D \citep{llava3d}, will average the features from different views, which might compromise the integrity of the information they convey. Furthest Point Sampling (FPS3D) will select points regularly in space, but the corresponding selected views will be random, which reduces the amount of information available to the LLM. 

To enable similar performances on a point-based structure, we propose to use a novel view-sensitive subsampling method. Our proposed FPS6D samples points regularly in space but also maximizes the diversity of selected views. As illustrated in \Cref{fig:method} (b), we use the FPS algorithm but in a higher-dimensional space that accounts for 3D positions and views. This is achieved by concatenating the point 3D positions and the 3D coordinates of the views from which they were seen. In this 6-dimensional space, the L2 distance between each point depends on both the distance in the 3D world and the distance between the 3D coordinates of the corresponding views. Note that we use the 3D camera positions to encode the views instead of a simple discrete view index. This is to help sample diverse viewpoints, even when two different views are close to each other.

Finally, the order in which the tokens are fed to the LLM also matters for the performance. The LLM backbone still uses sequence position encodings, which are different from the 3D position encodings we add to the unified spatial tokens. They allow the LLM to know where each token is located in the sequence. For the video-based structure, the order is already well defined as the order of patches in the video. For point based-structure, there is more room for customization as the points don't have a predefined order, and FPS6D returns points in a random order. We compare different token permutations in our experiments and opt for an object-based order, where the points inside the bounding box of each detected object (from the smallest to the biggest) are placed in groups, after the rest of the scene tokens, ordered based on their corresponding patch order in the video.

\subsection{Implementation details}

Our method builds upon the Video-3D LLM framework \citep{video3dllm}, adopting its multi-task training strategy and core implementation details for reproducibility. We train a single model on a diverse 3D scene understanding dataset, including 3D visual grounding, 3D dense captioning, and 3D question answering tasks. During training, we randomly sample a single task type per batch, optimizing the model exclusively on task-specific data within that batch. For 3D question answering and dense captioning, we employ cross-entropy loss to supervise text generation, while for 3D visual grounding, we use InfoNCE loss to optimize the selection of target objects from detected proposals, as detailed in \citep{video3dllm}. The model is based on the LLaVA-Video 7B architecture and is trained for one epoch using the Adam optimizer with a batch size of 16 and a warmup ratio of 0.03. All experiments are conducted on 8 H100-80G GPUs. For additional details on the training setup, hyperparameters, and data preprocessing, we refer readers to \citep{video3dllm}.

%
%

\section{Experiments}

\subsection{Experimental Setup}

\begin{table}[b]
\centering
\caption{Performance comparison with state-of-the-art 3D generalist methods. "Our avg" numbers are averaged over ten seeds. \textbf{Bold} is best. \underline{Underlined} is within 1\% of best.}
\label{tab:sota_comparison}
\resizebox{\textwidth}{!}{
\begin{tabular}{l|c|cc|cc|cc|cc|c}
\toprule
Method & All &\multicolumn{2}{c|}{ScanRefer} & \multicolumn{2}{c|}{Multi3DRefer} & \multicolumn{2}{c|}{Scan2Cap} & \multicolumn{2}{c|}{ScanQA} & SQA3D \\
& NS & Ac25 & Ac50 & F$_1$25 & F$_1$50 & B$_4$50 & C50 & C & EM & EM \\

\midrule
\multicolumn{11}{c}{\textit{Object-based Structure}} \\
\midrule

LEO \citep{leo}	 & 87.0	 & -	 & -	 & -	 & -	 & 38.2	 & 72.4	 & 101.4	 & 21.5	 & 50.0	\\
ChatScene        \citep{chat_scene}	 & 91.5	 & 55.5	 & 50.2	 & 57.1	 & 52.4	 & 36.3	 & 77.2	 & 87.7	 & 21.6	 & 54.6	\\

\midrule											
\multicolumn{11}{c}{\textit{Point-based Structure}} \\											
\midrule											
											
SceneLLM \citep{scenellm} 	 & 86.8	 & -	 & -	 & -	 & -	 & -	 & -	 & 80.0	 & 27.2	 & 53.6	\\
Grounded 3D-LLM \citep{grounded_3dllm} 	 & 80.8	 & 47.9	 & 44.1	 & 45.2	 & 40.6	 & 35.5	 & 70.6	 & 72.7	 & -	 & -	\\
PQ3D \citep{pq3d} 	 & 92.7	 & 57.0	 & 51.2	 & -	 & 50.1	 & 36.0	 & 80.3	 & -	 & -	 & 47.1	\\
3DLLaVA \citep{deng20253dllava} 	 & 91.8	 & -	 & -	 & -	 & -	 & 36.9	 & 78.8	 & 92.6	 & -	 & 54.5	\\
LLaVA-3D \citep{llava3d} 	 & 91.9	 & 54.1	 & 42.4	 & -	 & -	 & 41.1	 & 79.2	 & 91.7	 & 27.0	 & 55.6	\\
Our avg (point-based)	 & \underline{101.1} $\pm$0.3	 & 59.4	 & 52.5	 & \underline{58.6}	 & 53.1	 & 40.6	 & \underline{86.1}	 & 102.1	 & \underline{29.9}	 & \underline{59.8}	\\
Ours best (point-based)	 & \underline{101.5}	 & 59.7	 & 52.8	 & \underline{58.8}	 & 53.2	 & 40.7	 & \textbf{86.8}	 & 102.1	 & \underline{29.8}	 & \textbf{60.3}	\\
											
\midrule											
\multicolumn{11}{c}{\textit{Video-based Structure}} \\											
\midrule											
											
Vid3DLLM \citep{video3dllm} (claimed) 	 & 100.5	 & 58.1	 & 51.7	 & 58.0	 & 52.7	 & \textbf{42.4}	 & 83.8	 & 102.1	 & \textbf{30.1}	 & 58.6	\\
Vid3DLLM \citep{video3dllm} (released) 	 & 100.0	 & 58.2	 & 51.8	 & 57.4	 & 52.1	 & 41.3	 & 83.9	 & 102.0	 & \underline{30.0}	 & 58.5	\\
Vid3DLLM \citep{video3dllm} (reprod) 	 & 99.5 $\pm$0.5	 & 58.5	 & 52.0	 & 57.6	 & 52.3	 & 40.2	 & 81.1	 & \underline{103.2}	 & 29.5	 & 59.0	\\
Our avg (video-based)	 & \underline{101.1} $\pm$0.6	 & \underline{60.2}	 & \underline{53.5}	 & \underline{58.7}	 & \underline{53.4}	 & 40.7	 & 83.0	 & \underline{103.4}	 & 29.6	 & 59.2	\\
Our best (video-based)	 & \textbf{101.8}	 & \textbf{60.6}	 & \textbf{53.9}	 & \textbf{59.1}	 & \textbf{53.8}	 & 41.5	 & 83.7	 & \textbf{103.7}	 & 29.5	 & 59.6	\\

\bottomrule
\end{tabular}
}
\end{table}

\textbf{Datasets.} We evaluate our model on five widely-used 3D scene understanding benchmarks, all derived from the ScanNet dataset \citep{dai2017scannet}, which includes 1,513 richly annotated RGB-D video scans of indoor scenes. For 3D visual grounding, we use ScanRefer \citep{chen2020scanrefer} for single-target object localization and Multi3DRefer \citep{zhang2023multi3drefer} for multi-target scenarios. For 3D dense captioning, we employ Scan2Cap \citep{chen2021scan2cap} to generate detailed object descriptions in 3D scenes. For 3D question answering, we utilize ScanQA \citep{azuma2022scanqa} for spatial reasoning tasks and SQA3D \citep{ma2022sqa3d} for situated reasoning. We preprocess RGB-D video frames at 3 FPS, extracting corresponding camera intrinsic and extrinsic parameters, following the protocol in \citep{video3dllm}. Evaluations are conducted on the validation sets for ScanRefer, Multi3DRefer, Scan2Cap, and ScanQA, and the test set for SQA3D, consistent with prior work \citep{chat_scene, leo, llava3d}.

\textbf{Metrics.} We adopt standard evaluation metrics for each benchmark. For ScanRefer, we report accuracy at IoU thresholds of 0.25 and 0.5 (Ac25, Ac50), where a prediction is correct if its IoU with the ground truth exceeds the threshold. For Multi3DRefer, we use F1 scores at IoU thresholds of 0.25 and 0.5 (F$_1$25, F$_1$50) to account for variable numbers of target objects. For Scan2Cap, we compute CIDEr@0.5IoU (C50) and BLEU-4@0.5IoU (B$_4$50), combining captioning metrics with IoU-based bounding box alignment. For ScanQA, we report CIDEr (C) and exact match accuracy (EM) to evaluate spatial reasoning. For SQA3D, we use exact match accuracy (EM) to assess situated reasoning performance. Finally, we also use a normalized score (NS) metric to get a sense of performance at a glance. We compute it as an average of multiple metrics relative to the state of the art. We use the Video-3D LLM released model scores $\left\{s^\mathrm{sota}_\text{Ac25}, \dots, s^\mathrm{sota}_\text{EM}, \right\}$ as the base scores to compute the normalized score:

\begin{equation}
s_\text{NS} = \frac{1}{|\text{M}|} \sum_{m \in \text{M}}{100 \times \frac{s_m}{s^\mathrm{sota}_m} } .
\end{equation}

During our ablation studies, we noticed trends, as some changes affected multiple metrics similarly, especially when they were from the same task. Therefore, to help with clarity, we use the normalized score for each task and the overall normalized score in our ablation studies. For 3DVG, $\text{M}=\left\{\text{Ac25}, \text{Ac50}, \text{F$_1$25}, \text{F$_1$50} \right\}$, for 3DCap  $\text{M}=\left\{\text{B$_4$50}, \text{C50} \right\}$ and for 3DQA $\text{M}=\left\{\text{C}, \text{EM}_\text{ScanQA}, \text{EM}_\text{SQA3D} \right\}$.

\subsection{Comparison with State-of-the-Art Methods}

We compare our approach with state-of-the-art  3D LLMs methods, as shown in \Cref{tab:sota_comparison}. For both our video-based and point-based structures, we conducted ten experiments with different random seeds to obtain an average performance, and we also show the results of the best model (highest NS value). Our point-based approach uses 8192 sampled points. For fairness, we use Mask3D-generated object proposals \citep{schult2023mask3d} for 3D visual grounding and dense captioning, consistent with prior work \citep{leo, chat_scene, video3dllm}. On average, both our point-based and video-based models outperform the state of the art, and our best video-based model sets a new standard, outperforming the released Video-3D LLM \citep{video3dllm} model by 2.4\% on ScanRefer Ac25, 1.7\% on Multi3DRefer F$_1$25, and 1.7\% on ScanQA C, among others. Our best point-based model improves SoTA on Scan2Cap C50 by 2.9\% and on SQA3D EM by 1.8\%.

\subsection{Influence of the 3D Encoder Integration}
\label{sec:exp_3denc}

To assess the impact of integrating explicit 3D geometric features, we conduct an ablation study on different strategies for incorporating the Sonata point cloud encoder. This experiment uses the video-based token structure for all variants to ensure a direct and fair comparison with the Video-3D LLM baseline, which relies solely on image features (img) and positional encodings (PE). Our baseline, denoted as 'img+PE', replicates the Video-3D LLM setup. We then explore several ways to incorporate the Sonata features (3D). We can directly add the 3D features without any projection (we repeat the features to get to the token dimension). This can be done with (\emojifire) or without (\emojifreeze) fine-tuning the PTv3 model. Instead of fine-tuning the full model, we can also keep it frozen and use a learned projector; either a linear layer or a 2-layer perceptron (MLP). 

The results in \Cref{tab:ablation_3denc} demonstrate the benefits of incorporating 3D features, particularly for the 3DVG and 3DCap tasks. Introducing a trainable linear projector (img+PE+3D\emojifreeze+linear) yields the best overall performance, with a 1.6\% NS improvement over the baseline. While the 2-layer MLP and the fully fine-tuned model show competitive results, they do not consistently outperform the simpler linear projector, which also has the advantage of reduced computational cost.

\begin{table}[ht]
\centering
\caption{Effect of different 3D-encoder integration strategies. All methods use the video-based token structure. Numbers are averaged over five seeds. \textbf{Bold} is best. \underline{Underlined} is within 1\% of best.}
\label{tab:ablation_3denc}
\resizebox{\textwidth}{!}{
\begin{tabular}{l|c|c|c|c}
\toprule

Method & NS (All) & NS (3DVG) & NS (3DCap) & NS (3DQA)  \\

\midrule

img+PE	 & 99.5 ($\pm$0.52)	 & 100.4 ($\pm$0.59)	 & 97.1 ($\pm$1.66)	 & \underline{100.1} ($\pm$0.56)	\\
img+PE+3D\emojifreeze	 & 99.7 ($\pm$1.08)	 & 100.8 ($\pm$1.34)	 & 96.5 ($\pm$2.52)	 & \underline{100.4} ($\pm$0.82)	\\
img+PE+3D\emojifire	 & \underline{100.5} ($\pm$0.50)	 & 101.8 ($\pm$0.65)	 & 97.6 ($\pm$2.35)	 & \textbf{100.7} ($\pm$0.76)	\\
img+PE+3D\emojifreeze+MLP	 & \underline{100.6} ($\pm$0.63)	 & 101.8 ($\pm$0.62)	 & \underline{98.7} ($\pm$1.21)	 & \underline{100.2} ($\pm$0.60)	\\
img+PE+3D\emojifreeze+linear	 & \textbf{101.1} ($\pm$0.56)	 & \textbf{102.8} ($\pm$0.70)	 & \textbf{98.8} ($\pm$1.52)	 & \underline{100.3} ($\pm$0.56)	\\

\bottomrule
\end{tabular}
}
\end{table}

\subsection{Token-Structure Variants}
\label{sec:exp_token_variants}

We now compare the different token structures mentioned above. First, we test three different subsampling algorithms to create a point-based structure: a grid-voxel averaging method similar to LLaVA-3D \citep{llava3d}, a standard FPS in 3D space, and our FPS in 6D space. The grid-based structure provides variable sequence length (average $\sim$2000 tokens). Both FPS methods are set to sample 4096 tokens. We also compare to our best video-based structure that uses a sequence of 32 views,  accounting for 6272 tokens. Note that we study the impact of the number of tokens in \Cref{sec:exp_fps6d}.

The results presented in \Cref{tab:ablation_token_struct} highlight the superiority of our FPS6D approach over other standard subsampling algorithms. This validates the importance of our strategy that balances both spatial coverage within the 3D scene and viewpoint diversity, ensuring that the selected tokens capture information from a varied set of perspectives. Finally, while the video-based structure still achieves the best overall results, our FPS6D point-based approach demonstrates strong competitiveness, surpassing the video-based structure on the 3DCap task, which we believe is the task that benefits the most from the token permutation based on object bounding boxes.

\begin{table}[h]
\centering
\caption{Effect of different token structures.  All methods use 32 views. Numbers are averaged over five seeds. \textbf{Bold} is best. \underline{Underlined} is within 1\% of best.}
\label{tab:ablation_token_struct}
\resizebox{\textwidth}{!}{
\begin{tabular}{l|c|c|c|c}
\toprule
Method & NS (All) & NS (3DVG) & NS (3DCap) & NS (3DQA)  \\

\midrule

Avg. Grid. (0.2m)	 & 98.6 ($\pm$0.27)	 & 100.0 ($\pm$0.37)	 & 96.5 ($\pm$0.83)	 & 98.2 ($\pm$0.98)	\\
FPS3D (4096 pts)   	 & 99.4 ($\pm$0.23)	 & 100.8 ($\pm$0.08)	 & 97.9 ($\pm$1.11)	 & 98.4 ($\pm$0.05)	\\
FPS6D (4096 pts)   	 & \underline{100.5} ($\pm$0.31)	 & 101.4 ($\pm$0.63)	 & \textbf{100.0} ($\pm$0.74)	 & \underline{99.8} ($\pm$0.17)	\\
\midrule					
Video (32 views)	 & \textbf{101.1} ($\pm$0.56)	 & \textbf{102.8} ($\pm$0.70)	 & 98.8 ($\pm$1.52)	 & \textbf{100.3} ($\pm$0.56)	\\

\bottomrule
\end{tabular}}
\end{table}

\subsection{Influence of Token Permutation}
\label{sec:exp_token_order}

In this experiment, we confirm that the token order affects the performance by testing different permutations. First, with the video-based structure, we verify that the LLM sequence position encodings are effectively being used by the LLM backbone. Indeed, as shown in \Cref{tab:ablation_order}, the standard \textit{patch} order outperforms \textit{random} permutations at every forward pass.

Then we compare several reasonable permutations for the point-based structure. First, we confirm that the \textit{default} order returned by FPS6D performs similarly to \textit{random} permutations at every forward pass. Then we observe a boost in performance when tokens are ordered based on their original \textit{patch} position, and an even better performance when the tokens are grouped by \textit{objects} and moved to the end of the sequence after the remaining scene tokens. Notably, we see a relatively high 3.4\% performance increase for the 3DCap task when using the \textit{objects} permutation. Finally, we note that stable permutations (\textit{patch} or \textit{objects}) not only improve the performance but also reduce the results variance, confirming that the LLM backbone needs structured sequences of tokens for stability.

\begin{table}[h]
\centering
\caption{Effect of different token permutations for point-based and video-based structures. Numbers are averaged over five seeds. \textbf{Bold} is best. \underline{Underlined} is within 1\% of best.}
\label{tab:ablation_order}
\resizebox{\textwidth}{!}{
\begin{tabular}{l|c|c|c|c}
\toprule
Struct-Permutation & NS (All) & NS (3DVG) & NS (3DCap) & NS (3DQA)  \\

\midrule

Point-\it{objects}	 & \underline{100.5} ($\pm$0.31)	 & 101.4 ($\pm$0.63)	 & \textbf{100.0} ($\pm$0.74)	 & \underline{99.8} ($\pm$0.17)	\\
Point-\it{patch}	 & 98.9 ($\pm$0.41)	 & 99.4 ($\pm$0.36)	 & 96.6 ($\pm$0.16)	 & \underline{99.7} ($\pm$0.77)	\\
Point-\it{default}	 & 95.3 ($\pm$1.08)	 & 97.5 ($\pm$0.42)	 & 89.2 ($\pm$3.14)	 & 96.4 ($\pm$0.72)	\\
Point-\it{random}	 & 94.3 ($\pm$2.03)	 & 97.5 ($\pm$0.10)	 & 84.9 ($\pm$8.36)	 & 96.2 ($\pm$0.70)	\\
\midrule					
Video-\it{patch}	 & \textbf{101.1} ($\pm$0.56)	 & \textbf{102.8} ($\pm$0.70)	 & 98.8 ($\pm$1.52)	 & \textbf{100.3} ($\pm$0.56)	\\
Video-\it{random}	 & 97.0 ($\pm$0.70)	 & 99.4 ($\pm$0.18)	 & 93.5 ($\pm$1.83)	 & 96.2 ($\pm$0.69)	\\

\bottomrule
\end{tabular}
}
\end{table}

\subsection{Number of Views and Subsampled Points}
\label{sec:exp_fps6d}

Finally, we examine the efficiency/accuracy trade-off by varying (i) the number of camera views for the video-based token structure and (ii) the number of sampled points for the FPS6D point-based structure. 
Results and inference times (ScanQA validation, single H100 GPU) are presented in \Cref{tab:views_points_ablation}. Increasing the number of source views generally improves performance for both structures. However, for point-based methods with a fixed, smaller token budget (e.g., 1024-2048 points), performance gains saturate when increasing source views beyond 24, as the limited tokens cannot fully leverage the added view diversity. For these point-based methods, each additional 8 source views adds approximately 50ms to inference time, primarily reflecting the SigLIP processing cost for the new views before 3D feature integration. For the point-based structure, increasing the number of sampled points consistently improves accuracy but also increases inference time due to more tokens being fed to the LLM. The configuration using 8192 sampled points from 32 views achieves a top average NS of 101.1 (matching the 32-view video-based approach), but is also the slowest (646ms). When comparing structures with the same number of source views, video-based generally outperforms point-based if the latter uses a comparable or slightly larger number of tokens (e.g., with 16 source views, video-based with 3136 tokens achieves NS 97.9, while point-based with 4096 tokens scores NS 97.4). However, with a large token budget (8192 points) and ample views (32), our point-based approach is on par with the video-based one (NS 101.1). Point-based methods incur higher initial feature extraction costs due to the 3D encoder, but as token counts rise significantly for either method, the LLM inference time becomes the more dominant component of the total inference time.

\begin{table}[h]
\centering
\caption{Efficiency and accuracy trade-off. We vary the number of source views and the number of sampled points for the point-based structure. The amount of tokens for the point-based structure is equal to the number of sampled points. The amount of tokens for the video-based structure is equal to the number of patches. Results are averaged over three seeds. \textbf{Bold} is best. \underline{Underlined} is within 1\% of best.}
\label{tab:views_points_ablation}
\resizebox{\textwidth}{!}{%
\begin{tabular}{lc|cc|cc|cc|cc}

\toprule
 \multicolumn{2}{c|}{\#Views:}  & \multicolumn{2}{c|}{8} & \multicolumn{2}{c|}{16} & \multicolumn{2}{c|}{24} & \multicolumn{2}{c}{32} \\
 \multicolumn{2}{c|}{\#Patches:} & \multicolumn{2}{c|}{1568} & \multicolumn{2}{c|}{3136} & \multicolumn{2}{c|}{4704} & \multicolumn{2}{c}{6272} \\

\midrule
 Structure & \#Tokens & NS (All) & $t_\text{infer}$  & NS (All) & $t_\text{infer}$ & NS (All) & $t_\text{infer}$ & NS (All) & $t_\text{infer}$  \\
\midrule

Point-based & 1024	 & 89.0 ($\pm$0.73) & 275ms	 & 95.2 ($\pm$0.86) & 327ms	 & 96.8 ($\pm$0.23) & 373ms	 & 96.8 ($\pm$0.31) & 427ms	\\
Point-based & 2048	 & 90.4 ($\pm$0.38) & 299ms	 & 96.6 ($\pm$0.32) & 350ms	 & 98.6 ($\pm$0.21) & 398ms	 & 98.6 ($\pm$0.14) & 449ms	\\
Point-based & 4096	 & 91.1 ($\pm$0.25) & 361ms	 & 97.4 ($\pm$0.61) & 412ms	 & 99.8 ($\pm$0.36) & 455ms	 & \underline{100.5} ($\pm$0.46) & 515ms	\\
Point-based & 8192	 & 91.1 ($\pm$0.45) & 498ms	 & 97.8 ($\pm$0.53) & 547ms	 & \underline{100.1} ($\pm$0.32) & 595ms	 & \textbf{101.1} ($\pm$0.34) & 646ms	\\
\midrule					
Video-based & \#Patches	 & 90.5 ($\pm$0.53) & 252ms	 & 97.9 ($\pm$0.66) & 341ms	 & \underline{100.4} ($\pm$0.58) & 451ms	 & \textbf{101.1} ($\pm$0.56) & 568ms	\\

\bottomrule
\end{tabular}%
}
\vspace{-3ex}
\end{table}


\section{Conclusion}

This work investigated 3D token structures for Multimodal Large Language Models (MLLMs), demonstrating that fusing explicit 3D geometric features from a Sonata encoder with 2D image semantics significantly boosts performance. We systematically compared video-based and point-based tokenization, introducing FPS6D, a novel view-sensitive sampling method that allows point-based structures to achieve highly competitive results. Our approach, validated through multi-seed averaging, sets new state-of-the-art benchmarks in 3D scene understanding.

Despite these advancements, limitations remain. Firstly, the integration of the Sonata encoder and richer token sequences increases computational overhead, potentially hindering real-time applications. Secondly, the MLLM backbone, having been pretrained extensively on 2D video data, possesses an innate affinity for video-based token sequences. This could inadvertently lead to an underestimation of point-based structures, whose full potential might only be unlocked with MLLMs more deeply grounded in 3D-native data through dedicated pretraining or adaptation on large-scale 3D datasets. Thirdly, while we explored scene-level tokenization, the vast design space, including object-centric or adaptive tokenization strategies, offers avenues for future exploration.

Addressing these limitations will be crucial for developing more versatile and efficient 3D-aware MLLMs. We believe our analysis provides a solid foundation for continued progress in this domain.


\bibliographystyle{unsrtnat}
\bibliography{IEEEabrv, main}

\newpage
\appendix
\setcounter{table}{0}   
\setcounter{figure}{0}  
\renewcommand{\thetable}{\Alph{table}}     
\renewcommand{\thefigure}{\Alph{figure}}   

\begin{center}
    \textbf{\Huge Appendix}
    \vspace{5ex}
\end{center}

%
%
\begin{abstract}
This supplementary material provides additional details and results to complement our main paper.
Section \ref{sec:fps6d_warping} offers an in-depth exploration of the view-sensitive Farthest Point Sampling algorithm adapted using 6D inputs (FPS6D), including its formal definition, pseudocode, and a study on a weighting parameter $w$ that balances the influence of 3D spatial coordinates and camera view positions. We discuss the impact of this parameter on point cloud statistics and provide visualizations.
Section \ref{sec:complete_results} reports the complete results for all experiments conducted in the main paper, including the results from the multiple seeds used, and all metrics.
Finally, Section \ref{sec:code} provides information regarding the open-source availability of our code to facilitate reproducibility and further research.
\end{abstract}

%
\section{FPS6D and Space Warping}
\label{sec:fps6d_warping}

In this section, we provide a detailed description of the Farthest Point Sampling (FPS) algorithm, its extension to FPS6D used in our work for point cloud tokenization, and an analysis of how a weighting parameter $w$ can be used to modulate the sampling behavior by warping the joint 3D-view space.

\subsection{Farthest Point Sampling (FPS)}

Farthest Point Sampling is an iterative algorithm widely used to sample a subset of points from a larger point set, such that the sampled points are maximally distant from each other, providing good coverage of the entire set.

Let $\mathcal{P} = \{p_1, p_2, \dots, p_N\}$ be a set of $N$ input points in a $D$-dimensional space, and let $M$ be the desired number of points to sample ($M \le N$). FPS aims to select a subset $\mathcal{S} = \{s_1, s_2, \dots, s_M\} \subseteq \mathcal{P}$ that are well-spread. 

The algorithm first initializes $\mathcal{S}$ by selecting an arbitrary point from $\mathcal{P}$ (e.g., the first point, or a random point) as $s_1$. Then, it iteratively selects the point that is the farthest from the current set $\mathcal{S}$. For each point $p_i \in \mathcal{P} \setminus \mathcal{S}$, we compute its minimum distance to any point already in $\mathcal{S}$: $d(p_i, \mathcal{S}) = \min_{s_k \in \mathcal{S}} \lVert p_i - s_k \rVert_2$. Then we select the point that maximizes this minimum distance $s_j = \arg\max_{p_i \in \mathcal{P} \setminus \mathcal{S}} d(p_i, \mathcal{S})$, and add it to $\mathcal{S}$. A pseudocode representation of the FPS algorithm is provided in Algorithm \ref{alg:fps}. Note that for efficiency, we use an optimized version of FPS proposed in \citep{han2023quickfps}.

\begin{algorithm}[H]
\caption{Farthest Point Sampling (FPS)}
\label{alg:fps}
\begin{algorithmic}[1]
\Require Set of $N$ points $\mathcal{P} = \{p_1, \dots, p_N\}$, number of points to sample $M$.
\Ensure Subset of $M$ points $\mathcal{S} = \{s_1, \dots, s_M\}$.
\State Initialize $\mathcal{S} \leftarrow \emptyset$.
\State Let $s_1$ be a randomly chosen point from $\mathcal{P}$ (or $p_1$).
\State Add $s_1$ to $\mathcal{S}$.
\State Initialize $D[i] \leftarrow \lVert p_i - s_1 \rVert_2^2$ for all $p_i \in \mathcal{P}$. \Comment{Squared Euclidean distance for efficiency}
\For{$j \leftarrow 2$ to $M$}
    \State Let $s_j$ be the point $p_k \in \mathcal{P}$ that maximizes $D[k]$.
    \State Add $s_j$ to $\mathcal{S}$.
    \For{each point $p_i \in \mathcal{P}$}
        \State $D[i] \leftarrow \min(D[i], \lVert p_i - s_j \rVert_2^2)$.
    \EndFor
\EndFor
\State \Return $\mathcal{S}$.
\end{algorithmic}
\end{algorithm}


\subsection{FPS6D: Adapting FPS for Joint 3D-View Information}

In our work, we utilize a variant of FPS, termed FPS6D, to sample points from a 3D scene represented by multiple views. For each point $p(k,i) \in \mathbb{R}^3$ ($i^{th}$ point viewed by camera $k$), and the 3D position of the camera $t_k \in \mathbb{R}^3$ from which it was observed, we define a 6-dimensional point representation:
\begin{equation}
\label{eq:fixed6d}
p_{\text{raw\_6D}}(k, i) = \begin{bmatrix} p(k, i) \\ t_k \end{bmatrix} \in \mathbb{R}^6.
\end{equation}
FPS6D then applies the standard FPS algorithm (Algorithm \ref{alg:fps}) to this set of 6D points. The distance metric used is the squared Euclidean distance in this 6D space.
This formulation allows the sampling to consider both the spatial distribution of points in the 3D scene and the distribution of viewpoints from which these points are observed.
The 3D distance component encourages selecting points that are physically spread out in the scene. The view distance component encourages selecting points observed from diverse camera positions. These two types of distances have different properties and scales, which motivates the exploration of a weighting mechanism.


\subsection{The Weight Parameter $w$: Warping the 6D Space}
To control the relative importance of the 3D spatial distance with respect to the view distance, we introduce a weight parameter $w \in [0, 1]$. Formally, we define a new warped 6D vector for each point $i$ from view $k$ as:
\begin{equation}
\label{eq:warped6d}
p_\mathrm{6D}(k, i) = \begin{bmatrix}
\sqrt{1-w} \cdot p(k, i) \\
\sqrt{w} \cdot t_k
\end{bmatrix} \:.
\end{equation}

The parameter $w$ allows us to warp this joint space:
\begin{itemize}
    \item If $w=0$, the sampling is based solely on the 3D coordinates $p(k,i)$. FPS6D behaves like standard 3D FPS, selecting points that are spatially distant in the scene, potentially ignoring viewpoint diversity.
    \item If $w \rightarrow 1$, the points from each view are so far apart from each other that FPS6D behaves like standard 3D FPS applied to each view independently. Note that $w=1$ does not make sense because we only have a handful of views. As soon as a point has been sampled from each view, all minimum distances will be 0, and points will be selected randomly.
    \item If $0 < w < 1$, the sampling considers a combination of both spatial and view distances. The exact balance is determined by $w$ and the relative scales of spatial coordinates versus camera coordinates.
\end{itemize}

In the main paper, we used a fixed strategy for FPS6D equivalent to $w=0.5$, as defined in \cref{eq:fixed6d}. We chose not to tune or rely on $w$ for the main paper experiments because it introduces an unnecessary parameter with a limited impact on final results, as we show in the following. However, studying the effect of $w$ provides valuable insights into the interplay between spatial and view diversity in the sampling process, which we explore below.


\subsection{Impact of $w$ on Point Cloud Statistics}

The choice of $w$ directly influences the characteristics of the sampled point cloud tokens. To measure this impact, we use three metrics: 
\begin{itemize}
    \item \textbf{Number of points per view}: We measure the standard deviation (StdDev) of the total number of sampled points per view. This indicates if the views are sampled more regularly (lower StdDev) or more randomly (higher StdDev).
    \item \textbf{Number of views per neighborhood}: After subsampling, we count how many views are represented in the 32 nearest neighbors of each point (using 3D distance). A higher value indicates a larger diversity of sampled views in local regions.
    \item \textbf{Distance to closest point}: After subsampling, we measure the distance to the closest neighbor of each point (using 3D distance). This shows how uniform the spatial distribution of points is. A higher value and lower standard deviation indicate that points are spread more regularly in the 3D space.
\end{itemize}

\begin{figure}[h!]
  \centering
  \includegraphics[width=0.99\linewidth]{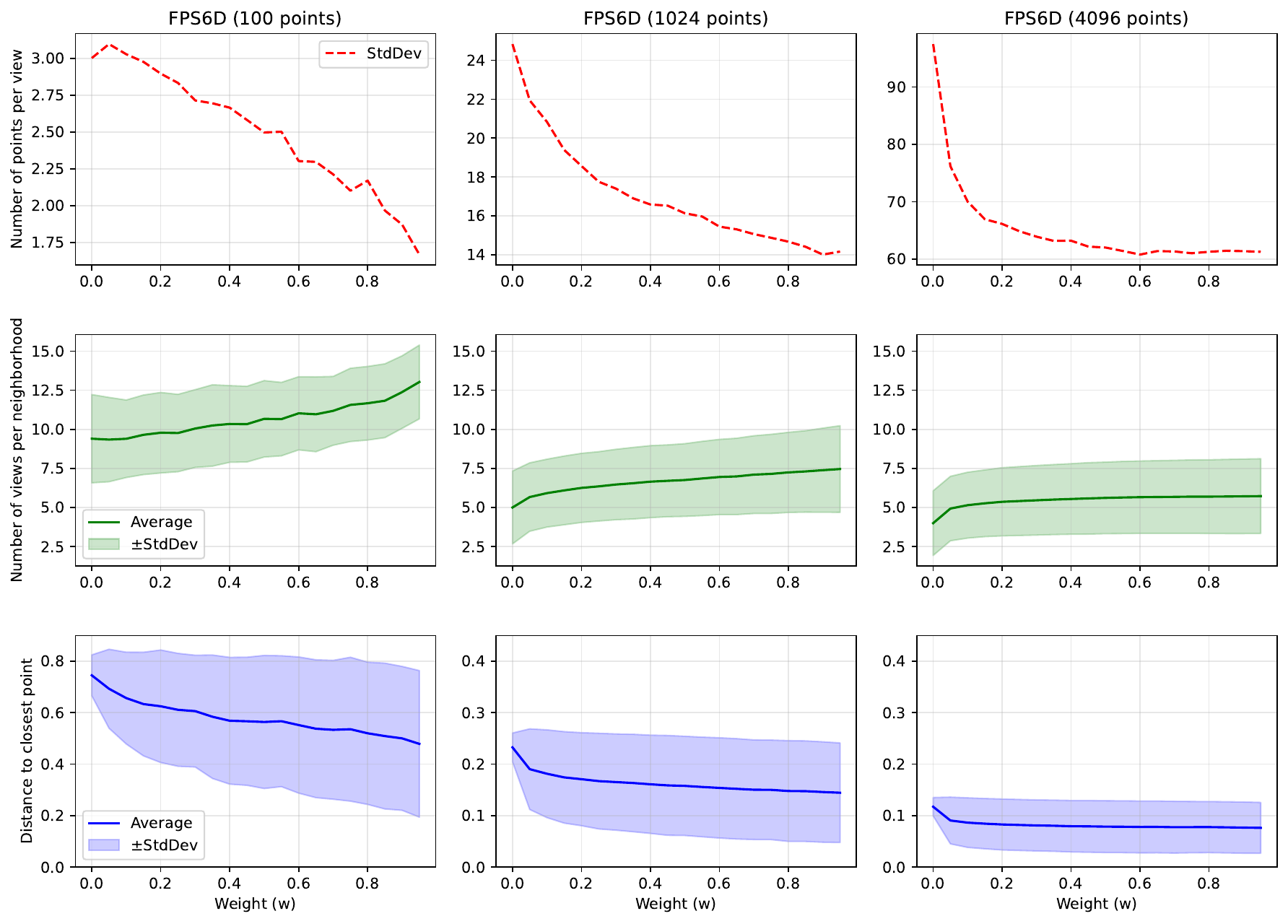}
  \caption{Influence of the weight parameter $w$ on spatial coverage and viewpoint selections. We show the variations of three metrics for different numbers of sampled points and multiple values of $w$. Metrics are averaged over multiple scenes.}
  \label{fig:w_influence_scenes}
\end{figure}

We visualize the effect of $w$ on the sampled points for representative scenes in Figure \ref{fig:w_influence_scenes}. Each line represents a metric, and each column represents a different number of points sampled by FPS6D. These visualizations show how different values of $w$ lead to different spatial coverage and viewpoint selections.

\begin{figure}[t]
  \centering
  \includegraphics[width=0.99\linewidth]{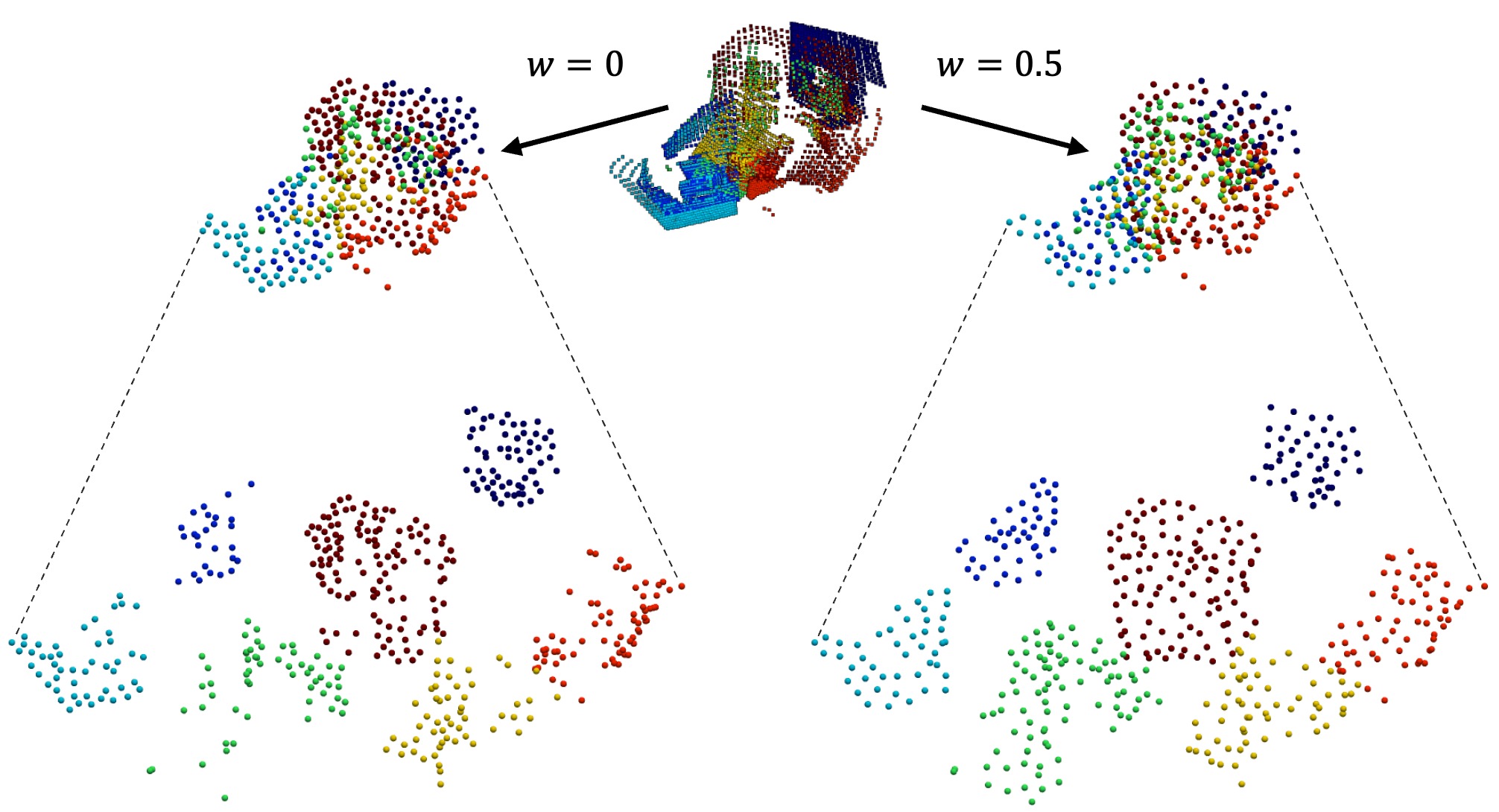}
  \caption{Visual comparison of points sampled by FPS6D with varying $w$ values on a sample scene. Points are colored by their original camera ID, and separated by camera ID at the bottom. We notice that the combined point cloud is more regular for $w=0$ (purely spatial sampling), but the point clouds from individual views are not as regularly sampled and are missing parts.}
  \label{fig:w_effect}
\end{figure}

For every column, we see that the most regular spatial distribution is achieved with $w=0$ at the cost of a high randomness of the selected view. When increasing $w$, the spatial regularity drops, and the view diversity improves. This is particularly visible for a lower number of sampled points. When sampling more points, we notice that the spatial coverage and view diversity do not change much for different values of $w$ except $w=0$. The FPS6D algorithm quickly shifts from a FPS3D regime to a stable regime with better view diversity but worse spatial distribution. Therefore, when sampling a larger number of points like we do in our experiments, it makes more sense to compare FPS3D and one version of FPS6D (for example $w=0.5$), which is why we did not introduce the $w$ parameter in the main paper. \Cref{fig:w_effect} illustrates the effect of $w$ on a real point cloud from one of the ScanNet scenes.

%
\section{Complete Results for All Experiments}
\label{sec:complete_results}

In the spirit of full transparency, we provide the results from all the experiments in the main paper. This also gives an idea of how spread the results are for each metric, in more detail than the standard deviations reported in the main paper. The results for all our ablations are compiled in \cref{tab:ablation_3denc_full,tab:token_structure_full,tab:token_permutation_full,tab:views_points_video_full,tab:views_points_point_full}

%
\section{Open-Source Code for Reproducing Results}
\label{sec:code}

To facilitate reproducibility and encourage further research in this area, we will make our code and pre-trained model checkpoints publicly available.
The repository will include: source code for preprocessing, training, test, and evaluations. We will also provide pre-trained weights for our best models. Our code will be available at: {\color{mypink}\url{https://github.com/apple/ml-pts3dllm}}.

%
%

\begin{table}[b]
\centering
\caption{Full results for our 3D encoder integration study (main paper table 2). Also used in the state-of-the-art (main paper table 1, video-based). \textbf{Bold} is best. \underline{Underlined} is within 1\% of best.
{\color{green2}Green} is better than SoTA.}
\label{tab:ablation_3denc_full}
\resizebox{\textwidth}{!}{
\begin{tabular}{l|c|cc|cc|cc|cc|c}
\toprule
Method & All &\multicolumn{2}{c|}{ScanRefer} & \multicolumn{2}{c|}{Multi3DRefer} & \multicolumn{2}{c|}{Scan2Cap} & \multicolumn{2}{c|}{ScanQA} & SQA3D \\
& NS & Ac25 & Ac50 & F$_1$25 & F$_1$50 & B$_4$50 & C50 & C & EM & EM \\

\midrule
\multirow{6}{*}{img+PE} & {\color{grey}98.6} & {\color{green2}58.3} & {\color{green2}51.9} & {\color{grey}57.2} & {\color{grey}52.0} & {\color{grey}39.2} & {\color{grey}77.6} & {\color{green2}102.9} & {\color{grey}29.7} & {\color{green2}59.2} \\
 & {\color{grey}99.2} & {\color{grey}57.9} & {\color{grey}51.4} & {\color{green2}57.5} & {\color{green2}52.3} & {\color{grey}40.1} & {\color{grey}81.3} & {\color{green2}102.4} & {\color{grey}29.3} & {\color{green2}59.4} \\
 & {\color{grey}99.8} & {\color{green2}58.7} & {\color{green2}52.1} & {\color{green2}57.5} & {\color{green2}52.2} & {\color{grey}40.1} & {\color{grey}80.8} & {\color{green2}\underline{104.6}} & {\color{grey}30.0} & {\color{green2}59.0} \\
 & {\color{green2}100.1} & {\color{green2}59.2} & {\color{green2}52.6} & {\color{green2}58.1} & {\color{green2}52.8} & {\color{grey}40.7} & {\color{grey}81.9} & {\color{green2}103.1} & {\color{grey}29.6} & {\color{grey}58.4} \\
 & {\color{grey}99.5} & {\color{green2}58.2} & {\color{grey}51.7} & {\color{green2}57.4} & {\color{grey}52.1} & {\color{grey}40.9} & {\color{grey}82.9} & {\color{green2}103.1} & {\color{grey}29.0} & {\color{green2}58.8} \\
 & {\color{green2}100.1} & {\color{green2}58.9} & {\color{green2}52.4} & {\color{green2}57.7} & {\color{green2}52.6} & {\color{grey}40.3} & {\color{grey}81.9} & {\color{green2}103.4} & {\color{grey}29.7} & {\color{green2}59.5} \\
\midrule
\multirow{11}{*}{img+PE+3D\emojifreeze} & {\color{grey}99.3} & {\color{grey}58.2} & {\color{grey}51.7} & {\color{green2}57.6} & {\color{green2}52.5} & {\color{grey}40.3} & {\color{grey}81.8} & {\color{green2}102.7} & {\color{grey}29.2} & {\color{green2}58.7} \\
 & {\color{grey}99.7} & {\color{grey}58.0} & {\color{grey}51.3} & {\color{green2}57.4} & {\color{grey}52.1} & {\color{grey}40.3} & {\color{grey}82.5} & {\color{green2}\underline{104.3}} & {\color{grey}29.9} & {\color{green2}59.2} \\
 & {\color{grey}99.3} & {\color{green2}58.8} & {\color{green2}52.1} & {\color{grey}57.3} & {\color{grey}51.9} & {\color{grey}39.7} & {\color{grey}80.4} & {\color{green2}103.7} & {\color{grey}29.8} & {\color{green2}58.8} \\
 & {\color{grey}98.8} & {\color{grey}57.9} & {\color{grey}51.4} & {\color{grey}57.1} & {\color{grey}51.9} & {\color{grey}39.8} & {\color{grey}80.4} & {\color{green2}102.3} & {\color{grey}29.6} & {\color{green2}59.0} \\
 & {\color{grey}99.0} & {\color{grey}58.1} & {\color{grey}51.5} & {\color{green2}57.6} & {\color{green2}52.3} & {\color{grey}39.2} & {\color{grey}79.3} & {\color{green2}103.7} & {\color{grey}29.9} & {\color{green2}59.0} \\
 & {\color{green2}\underline{100.9}} & {\color{green2}58.5} & {\color{green2}51.9} & {\color{green2}57.8} & {\color{green2}52.5} & {\color{grey}40.8} & {\color{green2}83.9} & {\color{green2}\textbf{105.3}} & {\color{green2}\textbf{30.5}} & {\color{green2}\underline{59.9}} \\
 & {\color{grey}97.5} & {\color{grey}57.9} & {\color{grey}51.4} & {\color{grey}57.1} & {\color{grey}51.9} & {\color{grey}38.3} & {\color{grey}73.0} & {\color{green2}102.7} & {\color{grey}29.5} & {\color{green2}\underline{59.7}} \\
 & {\color{grey}99.5} & {\color{green2}58.8} & {\color{green2}52.3} & {\color{green2}57.8} & {\color{green2}52.5} & {\color{grey}39.8} & {\color{grey}79.5} & {\color{green2}103.5} & {\color{grey}29.5} & {\color{green2}59.2} \\
 & {\color{green2}\underline{101.0}} & {\color{green2}59.7} & {\color{green2}52.9} & {\color{green2}58.2} & {\color{green2}53.0} & {\color{grey}40.6} & {\color{grey}82.4} & {\color{green2}\underline{104.6}} & {\color{grey}30.1} & {\color{green2}59.6} \\
 & {\color{green2}100.3} & {\color{green2}59.5} & {\color{green2}52.8} & {\color{green2}\underline{59.0}} & {\color{green2}\underline{53.6}} & {\color{grey}39.9} & {\color{grey}81.7} & {\color{grey}101.9} & {\color{grey}29.5} & {\color{green2}59.1} \\
 & {\color{green2}\underline{101.5}} & {\color{green2}\underline{60.7}} & {\color{green2}\underline{53.9}} & {\color{green2}\textbf{59.5}} & {\color{green2}\textbf{53.9}} & {\color{grey}40.7} & {\color{grey}83.3} & {\color{grey}101.9} & {\color{grey}29.8} & {\color{green2}59.3} \\
\midrule
\multirow{5}{*}{img+PE+3D\emojifire} & {\color{green2}\underline{100.9}} & {\color{green2}59.8} & {\color{green2}53.1} & {\color{green2}58.6} & {\color{green2}53.3} & {\color{grey}40.4} & {\color{grey}80.8} & {\color{green2}104.1} & {\color{grey}30.0} & {\color{green2}\textbf{60.3}} \\
 & {\color{green2}100.3} & {\color{green2}59.0} & {\color{green2}52.5} & {\color{green2}57.6} & {\color{green2}52.3} & {\color{grey}40.6} & {\color{grey}83.8} & {\color{green2}103.5} & {\color{grey}29.7} & {\color{green2}59.4} \\
 & {\color{grey}99.9} & {\color{green2}59.5} & {\color{green2}52.9} & {\color{green2}58.1} & {\color{green2}52.9} & {\color{grey}39.1} & {\color{grey}77.5} & {\color{green2}103.9} & {\color{green2}\underline{30.4}} & {\color{green2}\underline{59.8}} \\
 & {\color{green2}100.1} & {\color{green2}59.3} & {\color{green2}52.7} & {\color{green2}57.9} & {\color{green2}52.6} & {\color{grey}40.1} & {\color{grey}82.3} & {\color{green2}103.3} & {\color{grey}29.5} & {\color{green2}59.1} \\
 & {\color{green2}\underline{101.2}} & {\color{green2}60.0} & {\color{green2}53.3} & {\color{green2}58.6} & {\color{green2}53.3} & {\color{grey}\underline{41.2}} & {\color{green2}\underline{84.9}} & {\color{green2}102.3} & {\color{grey}29.3} & {\color{green2}59.6} \\
\midrule
\multirow{5}{*}{\shortstack{img+PE+3D\emojifreeze\\+MLP}} & {\color{green2}100.3} & {\color{green2}59.5} & {\color{green2}52.7} & {\color{green2}58.4} & {\color{green2}53.1} & {\color{grey}40.4} & {\color{grey}82.3} & {\color{grey}102.0} & {\color{grey}29.4} & {\color{green2}59.4} \\
 & {\color{green2}100.2} & {\color{green2}59.1} & {\color{green2}52.6} & {\color{green2}58.0} & {\color{green2}52.8} & {\color{grey}40.0} & {\color{grey}81.6} & {\color{green2}\underline{104.5}} & {\color{grey}29.6} & {\color{green2}59.1} \\
 & {\color{grey}99.9} & {\color{green2}58.7} & {\color{green2}52.1} & {\color{green2}58.0} & {\color{green2}52.7} & {\color{grey}40.6} & {\color{grey}82.5} & {\color{green2}102.6} & {\color{grey}29.3} & {\color{green2}58.9} \\
 & {\color{green2}\underline{101.6}} & {\color{green2}60.0} & {\color{green2}53.4} & {\color{green2}58.6} & {\color{green2}53.1} & {\color{green2}\underline{41.5}} & {\color{green2}\underline{84.6}} & {\color{green2}103.6} & {\color{grey}29.8} & {\color{green2}\underline{60.1}} \\
 & {\color{green2}\underline{100.9}} & {\color{green2}60.1} & {\color{green2}53.3} & {\color{green2}58.4} & {\color{green2}53.1} & {\color{grey}40.8} & {\color{grey}83.6} & {\color{green2}103.6} & {\color{grey}29.8} & {\color{grey}58.3} \\
\midrule
\multirow{13}{*}{\shortstack{img+PE+3D\emojifreeze\\+linear}} & {\color{green2}100.1} & {\color{green2}59.5} & {\color{green2}53.1} & {\color{green2}58.5} & {\color{green2}53.2} & {\color{grey}39.7} & {\color{grey}80.6} & {\color{green2}103.0} & {\color{grey}29.4} & {\color{green2}59.3} \\
 & {\color{green2}\underline{101.2}} & {\color{green2}\textbf{60.9}} & {\color{green2}\textbf{54.1}} & {\color{green2}\underline{59.1}} & {\color{green2}\underline{53.8}} & {\color{grey}40.6} & {\color{grey}82.8} & {\color{green2}102.3} & {\color{grey}29.6} & {\color{green2}58.5} \\
 & {\color{green2}\textbf{101.8}} & {\color{green2}\underline{60.6}} & {\color{green2}\underline{53.9}} & {\color{green2}\underline{59.1}} & {\color{green2}\underline{53.8}} & {\color{green2}\underline{41.5}} & {\color{grey}83.7} & {\color{green2}103.7} & {\color{grey}29.5} & {\color{green2}59.6} \\
 & {\color{green2}\underline{101.7}} & {\color{green2}\underline{60.5}} & {\color{green2}53.5} & {\color{green2}\underline{59.2}} & {\color{green2}\underline{53.7}} & {\color{grey}41.0} & {\color{green2}\textbf{85.4}} & {\color{green2}103.2} & {\color{grey}29.8} & {\color{green2}59.2} \\
 & {\color{green2}100.2} & {\color{green2}59.7} & {\color{green2}53.1} & {\color{green2}58.5} & {\color{green2}53.2} & {\color{grey}39.9} & {\color{grey}79.2} & {\color{green2}104.1} & {\color{grey}29.9} & {\color{green2}59.1} \\
 & {\color{green2}\underline{101.4}} & {\color{green2}\underline{60.7}} & {\color{green2}\underline{53.9}} & {\color{green2}\underline{59.3}} & {\color{green2}\underline{53.8}} & {\color{grey}40.4} & {\color{grey}81.7} & {\color{green2}103.3} & {\color{grey}30.1} & {\color{green2}59.2} \\
 & {\color{green2}\underline{101.6}} & {\color{green2}\underline{60.4}} & {\color{green2}\underline{53.7}} & {\color{green2}\underline{59.1}} & {\color{green2}\underline{53.7}} & {\color{grey}40.7} & {\color{grey}83.0} & {\color{green2}\underline{104.3}} & {\color{grey}30.0} & {\color{green2}59.4} \\
 & {\color{green2}\underline{101.5}} & {\color{green2}\underline{60.6}} & {\color{green2}\underline{53.8}} & {\color{green2}\underline{59.2}} & {\color{green2}\underline{53.8}} & {\color{grey}40.7} & {\color{green2}84.1} & {\color{green2}102.9} & {\color{grey}29.6} & {\color{green2}59.2} \\
 & {\color{green2}100.7} & {\color{green2}59.9} & {\color{green2}53.2} & {\color{green2}58.3} & {\color{green2}53.1} & {\color{green2}\underline{41.3}} & {\color{green2}84.0} & {\color{green2}103.5} & {\color{grey}29.3} & {\color{grey}57.8} \\
 & {\color{green2}\underline{101.2}} & {\color{green2}60.1} & {\color{green2}53.4} & {\color{green2}58.7} & {\color{green2}\underline{53.4}} & {\color{grey}40.8} & {\color{grey}83.8} & {\color{green2}103.7} & {\color{grey}29.4} & {\color{green2}59.6} \\
 & {\color{green2}\underline{101.5}} & {\color{green2}60.0} & {\color{green2}53.3} & {\color{green2}58.6} & {\color{green2}53.1} & {\color{green2}\textbf{41.5}} & {\color{green2}84.4} & {\color{green2}103.6} & {\color{grey}29.8} & {\color{green2}59.6} \\
 & {\color{green2}100.3} & {\color{green2}59.4} & {\color{green2}52.8} & {\color{green2}58.3} & {\color{green2}53.0} & {\color{grey}40.6} & {\color{grey}82.5} & {\color{green2}102.6} & {\color{grey}29.4} & {\color{green2}58.9} \\
 & {\color{green2}\underline{100.9}} & {\color{green2}\underline{60.3}} & {\color{green2}53.4} & {\color{green2}57.7} & {\color{green2}52.4} & {\color{grey}41.0} & {\color{grey}83.2} & {\color{green2}103.4} & {\color{grey}29.8} & {\color{green2}59.6} \\
\bottomrule

\end{tabular}
}
\end{table}

%
%

\begin{table}[b]
\centering
\caption{Full results for our token structure study (main paper table 3). \textbf{Bold} is best. \underline{Underlined} is within 1\% of best.
{\color{green2}Green} is better than SoTA.}
\label{tab:token_structure_full}
\resizebox{\textwidth}{!}{
\begin{tabular}{l|c|cc|cc|cc|cc|c}
\toprule
Method & All &\multicolumn{2}{c|}{ScanRefer} & \multicolumn{2}{c|}{Multi3DRefer} & \multicolumn{2}{c|}{Scan2Cap} & \multicolumn{2}{c|}{ScanQA} & SQA3D \\
& NS & Ac25 & Ac50 & F$_1$25 & F$_1$50 & B$_4$50 & C50 & C & EM & EM \\

\midrule
\multirow{5}{*}{Avg. Grid. (0.2m)} & {\color{grey}98.9} & {\color{green2}58.8} & {\color{green2}52.2} & {\color{green2}57.5} & {\color{green2}52.1} & {\color{grey}39.6} & {\color{grey}80.1} & {\color{grey}100.2} & {\color{grey}29.1} & {\color{green2}59.5} \\
 & {\color{grey}98.2} & {\color{grey}57.8} & {\color{grey}51.3} & {\color{green2}57.5} & {\color{grey}52.1} & {\color{grey}39.8} & {\color{grey}82.9} & {\color{grey}100.3} & {\color{grey}28.4} & {\color{grey}57.3} \\
 & {\color{grey}98.8} & {\color{grey}58.1} & {\color{grey}51.5} & {\color{green2}57.6} & {\color{green2}52.2} & {\color{grey}39.5} & {\color{grey}81.0} & {\color{grey}101.5} & {\color{grey}29.2} & {\color{green2}59.0} \\
 & {\color{grey}99.0} & {\color{grey}57.9} & {\color{grey}51.5} & {\color{grey}57.3} & {\color{grey}52.0} & {\color{grey}39.9} & {\color{grey}81.2} & {\color{grey}101.5} & {\color{grey}29.3} & {\color{green2}59.6} \\
 & {\color{grey}98.3} & {\color{grey}57.4} & {\color{grey}50.9} & {\color{grey}57.1} & {\color{grey}51.8} & {\color{grey}39.9} & {\color{grey}80.9} & {\color{grey}100.7} & {\color{grey}29.2} & {\color{green2}58.7} \\
\midrule
\multirow{7}{*}{FPS3D (4096 pts)} & {\color{grey}99.2} & {\color{green2}59.3} & {\color{green2}52.5} & {\color{green2}57.6} & {\color{green2}52.2} & {\color{grey}39.5} & {\color{grey}82.2} & {\color{grey}100.1} & {\color{grey}28.9} & {\color{green2}59.1} \\
 & {\color{grey}99.6} & {\color{green2}58.6} & {\color{green2}51.9} & {\color{green2}\underline{58.0}} & {\color{green2}\underline{52.6}} & {\color{grey}40.4} & {\color{green2}84.0} & {\color{grey}101.5} & {\color{grey}29.0} & {\color{grey}58.3} \\
 & {\color{grey}99.5} & {\color{green2}58.3} & {\color{grey}51.6} & {\color{green2}57.8} & {\color{green2}52.5} & {\color{grey}40.2} & {\color{grey}83.1} & {\color{grey}100.8} & {\color{grey}29.4} & {\color{green2}59.2} \\
 & {\color{green2}\underline{100.1}} & {\color{green2}58.6} & {\color{green2}51.9} & {\color{green2}\underline{58.1}} & {\color{green2}\underline{52.8}} & {\color{grey}\underline{41.0}} & {\color{green2}84.3} & {\color{grey}101.2} & {\color{grey}29.1} & {\color{green2}59.6} \\
 & {\color{grey}99.3} & {\color{green2}58.6} & {\color{green2}52.0} & {\color{green2}57.7} & {\color{green2}52.4} & {\color{grey}40.1} & {\color{grey}82.3} & {\color{grey}100.8} & {\color{grey}28.9} & {\color{green2}59.3} \\
 & {\color{grey}99.8} & {\color{green2}58.9} & {\color{green2}52.2} & {\color{green2}\underline{58.2}} & {\color{green2}\underline{52.8}} & {\color{grey}40.3} & {\color{green2}83.9} & {\color{grey}100.9} & {\color{grey}29.2} & {\color{green2}58.7} \\
 & {\color{grey}99.5} & {\color{green2}58.9} & {\color{green2}52.1} & {\color{green2}57.7} & {\color{green2}52.5} & {\color{grey}40.8} & {\color{green2}84.1} & {\color{grey}100.2} & {\color{grey}28.9} & {\color{grey}58.4} \\
\midrule
\multirow{8}{*}{FPS6D (4096 pts) } & {\color{green2}\underline{100.5}} & {\color{green2}59.0} & {\color{green2}52.4} & {\color{green2}57.8} & {\color{green2}52.4} & {\color{green2}\textbf{41.4}} & {\color{green2}84.9} & {\color{green2}\underline{102.6}} & {\color{grey}29.3} & {\color{green2}59.2} \\
 & {\color{green2}\underline{100.2}} & {\color{green2}59.0} & {\color{green2}52.2} & {\color{green2}\underline{58.2}} & {\color{green2}\underline{52.8}} & {\color{grey}40.5} & {\color{green2}84.0} & {\color{grey}101.4} & {\color{grey}29.2} & {\color{green2}\underline{60.0}} \\
 & {\color{green2}\underline{100.9}} & {\color{green2}\textbf{60.2}} & {\color{green2}\textbf{53.5}} & {\color{green2}\underline{58.3}} & {\color{green2}\underline{52.8}} & {\color{grey}40.6} & {\color{green2}84.3} & {\color{green2}\textbf{102.8}} & {\color{grey}29.5} & {\color{green2}59.4} \\
 & {\color{grey}\underline{100.0}} & {\color{green2}58.9} & {\color{green2}52.0} & {\color{green2}57.9} & {\color{green2}52.3} & {\color{grey}40.4} & {\color{green2}84.0} & {\color{grey}101.2} & {\color{grey}29.4} & {\color{green2}59.8} \\
 & {\color{green2}\textbf{101.0}} & {\color{green2}58.9} & {\color{green2}52.3} & {\color{green2}\underline{58.5}} & {\color{green2}\underline{53.1}} & {\color{grey}\underline{41.1}} & {\color{green2}\textbf{86.9}} & {\color{grey}101.5} & {\color{grey}29.4} & {\color{green2}\underline{60.1}} \\
 & {\color{green2}\underline{100.6}} & {\color{green2}59.3} & {\color{green2}52.6} & {\color{green2}\textbf{58.5}} & {\color{green2}\textbf{53.1}} & {\color{grey}40.8} & {\color{green2}85.1} & {\color{grey}101.8} & {\color{grey}29.0} & {\color{green2}59.7} \\
 & {\color{green2}\underline{100.5}} & {\color{green2}59.1} & {\color{green2}52.4} & {\color{green2}57.8} & {\color{green2}52.5} & {\color{grey}40.7} & {\color{green2}\underline{86.0}} & {\color{grey}101.1} & {\color{grey}29.3} & {\color{green2}\underline{59.9}} \\
 & {\color{green2}\underline{100.6}} & {\color{green2}59.1} & {\color{green2}52.5} & {\color{green2}\underline{58.2}} & {\color{green2}\underline{52.9}} & {\color{grey}40.5} & {\color{green2}84.3} & {\color{grey}\underline{101.8}} & {\color{grey}29.4} & {\color{green2}\textbf{60.5}} \\
\bottomrule

\end{tabular}
}
\end{table}

%
%

\begin{table}[b]
\centering
\caption{Full results for our token permutation study (main paper table 4). \textbf{Bold} is best. \underline{Underlined} is within 1\% of best.
{\color{green2}Green} is better than SoTA.}
\label{tab:token_permutation_full}
\resizebox{\textwidth}{!}{
\begin{tabular}{l|c|cc|cc|cc|cc|c}
\toprule
Method & All &\multicolumn{2}{c|}{ScanRefer} & \multicolumn{2}{c|}{Multi3DRefer} & \multicolumn{2}{c|}{Scan2Cap} & \multicolumn{2}{c|}{ScanQA} & SQA3D \\
& NS & Ac25 & Ac50 & F$_1$25 & F$_1$50 & B$_4$50 & C50 & C & EM & EM \\

\midrule
\multirow{3}{*}{Point-\it{patch}} & {\color{grey}\underline{99.1}} & {\color{grey}\underline{57.9}} & {\color{grey}\underline{51.5}} & {\color{grey}\underline{57.3}} & {\color{grey}\underline{52.0}} & {\color{grey}39.9} & {\color{grey}81.2} & {\color{green2}\underline{102.4}} & {\color{grey}29.3} & {\color{green2}\underline{59.6}} \\
 & {\color{grey}98.3} & {\color{grey}57.4} & {\color{grey}50.9} & {\color{grey}\underline{57.1}} & {\color{grey}\underline{51.8}} & {\color{grey}39.9} & {\color{grey}80.9} & {\color{grey}100.7} & {\color{grey}29.2} & {\color{green2}58.7} \\
 & {\color{grey}\underline{99.3}} & {\color{green2}\underline{58.2}} & {\color{grey}\underline{51.5}} & {\color{grey}\underline{57.2}} & {\color{grey}\underline{52.1}} & {\color{grey}39.7} & {\color{grey}81.0} & {\color{green2}\textbf{102.6}} & {\color{grey}\underline{29.9}} & {\color{green2}\textbf{59.6}} \\
\midrule
\multirow{3}{*}{Point-\it{default}} & {\color{grey}95.4} & {\color{grey}56.9} & {\color{grey}50.5} & {\color{grey}56.2} & {\color{grey}51.0} & {\color{grey}37.7} & {\color{grey}74.0} & {\color{grey}97.1} & {\color{grey}28.1} & {\color{grey}57.9} \\
 & {\color{grey}96.6} & {\color{grey}57.0} & {\color{grey}50.5} & {\color{grey}56.3} & {\color{grey}51.1} & {\color{grey}39.0} & {\color{grey}76.2} & {\color{grey}98.6} & {\color{grey}28.7} & {\color{green2}58.6} \\
 & {\color{grey}93.9} & {\color{grey}56.1} & {\color{grey}49.8} & {\color{grey}56.0} & {\color{grey}50.9} & {\color{grey}36.8} & {\color{grey}68.0} & {\color{grey}97.6} & {\color{grey}28.1} & {\color{grey}57.7} \\
\midrule
\multirow{3}{*}{Point-\it{random}} & {\color{grey}96.1} & {\color{grey}56.4} & {\color{grey}50.0} & {\color{grey}56.4} & {\color{grey}51.2} & {\color{grey}38.8} & {\color{grey}75.4} & {\color{grey}98.1} & {\color{grey}28.5} & {\color{green2}58.8} \\
 & {\color{grey}95.3} & {\color{grey}56.7} & {\color{grey}50.3} & {\color{grey}56.0} & {\color{grey}50.9} & {\color{grey}37.8} & {\color{grey}73.5} & {\color{grey}97.3} & {\color{grey}28.2} & {\color{grey}58.2} \\
 & {\color{grey}91.5} & {\color{grey}56.7} & {\color{grey}50.3} & {\color{grey}56.4} & {\color{grey}51.1} & {\color{grey}33.9} & {\color{grey}53.9} & {\color{grey}96.5} & {\color{grey}28.0} & {\color{grey}57.6} \\
\midrule
\multirow{3}{*}{Video-\it{random}} & {\color{grey}96.4} & {\color{grey}\underline{57.8}} & {\color{grey}\underline{51.3}} & {\color{grey}\underline{57.0}} & {\color{grey}\underline{51.7}} & {\color{grey}38.4} & {\color{grey}75.8} & {\color{grey}97.2} & {\color{grey}28.1} & {\color{grey}57.9} \\
 & {\color{grey}98.0} & {\color{green2}\textbf{58.3}} & {\color{grey}\underline{51.6}} & {\color{grey}\underline{57.0}} & {\color{grey}\underline{51.8}} & {\color{grey}39.7} & {\color{grey}80.4} & {\color{grey}98.4} & {\color{grey}28.8} & {\color{grey}58.2} \\
 & {\color{grey}96.7} & {\color{grey}\underline{57.9}} & {\color{grey}51.2} & {\color{grey}\underline{57.2}} & {\color{grey}\underline{52.0}} & {\color{grey}38.7} & {\color{grey}77.1} & {\color{grey}96.4} & {\color{grey}27.8} & {\color{green2}58.6} \\
\bottomrule

\end{tabular}
}
\end{table}

%
%

\begin{table}[b]
\centering
\caption{Full results for our number of views and subsampled points study (main paper table 5). Only video-based models are shown here. \textbf{Bold} is best. \underline{Underlined} is within 1\% of best.
{\color{green2}Green} is better than SoTA.}
\label{tab:views_points_video_full}
\resizebox{\textwidth}{!}{
\begin{tabular}{l|c|cc|cc|cc|cc|c}
\toprule
Method & All &\multicolumn{2}{c|}{ScanRefer} & \multicolumn{2}{c|}{Multi3DRefer} & \multicolumn{2}{c|}{Scan2Cap} & \multicolumn{2}{c|}{ScanQA} & SQA3D \\
& NS & Ac25 & Ac50 & F$_1$25 & F$_1$50 & B$_4$50 & C50 & C & EM & EM \\

\midrule
\multirow{1}{*}{origin} & {\color{green2}100.0} & {\color{green2}58.2} & {\color{green2}51.8} & {\color{green2}57.4} & {\color{green2}52.1} & {\color{green2}\underline{41.3}} & {\color{green2}83.9} & {\color{green2}102.0} & {\color{green2}\underline{30.1}} & {\color{green2}58.5} \\
\midrule
\multirow{6}{*}{\shortstack{Video-based\\8 views}} & {\color{grey}90.6} & {\color{grey}51.5} & {\color{grey}46.1} & {\color{grey}51.2} & {\color{grey}47.0} & {\color{grey}38.5} & {\color{grey}70.7} & {\color{grey}95.3} & {\color{grey}27.3} & {\color{grey}57.1} \\
 & {\color{grey}89.8} & {\color{grey}51.1} & {\color{grey}45.2} & {\color{grey}51.0} & {\color{grey}46.4} & {\color{grey}37.3} & {\color{grey}68.4} & {\color{grey}96.4} & {\color{grey}27.6} & {\color{grey}57.0} \\
 & {\color{grey}91.2} & {\color{grey}51.6} & {\color{grey}46.0} & {\color{grey}51.6} & {\color{grey}47.2} & {\color{grey}38.6} & {\color{grey}72.5} & {\color{grey}96.9} & {\color{grey}27.7} & {\color{grey}56.2} \\
 & {\color{grey}89.8} & {\color{grey}51.2} & {\color{grey}45.3} & {\color{grey}50.8} & {\color{grey}46.2} & {\color{grey}37.8} & {\color{grey}69.5} & {\color{grey}96.0} & {\color{grey}27.3} & {\color{grey}56.3} \\
 & {\color{grey}90.9} & {\color{grey}51.6} & {\color{grey}45.9} & {\color{grey}50.8} & {\color{grey}46.2} & {\color{grey}38.3} & {\color{grey}73.2} & {\color{grey}95.9} & {\color{grey}27.6} & {\color{grey}57.4} \\
 & {\color{grey}90.7} & {\color{grey}51.8} & {\color{grey}45.9} & {\color{grey}50.7} & {\color{grey}46.1} & {\color{grey}38.3} & {\color{grey}72.7} & {\color{grey}95.8} & {\color{grey}27.8} & {\color{grey}56.3} \\
\midrule
\multirow{5}{*}{\shortstack{Video-based\\16 views}} & {\color{grey}96.9} & {\color{grey}57.3} & {\color{grey}50.9} & {\color{grey}56.4} & {\color{grey}51.3} & {\color{grey}38.8} & {\color{grey}77.2} & {\color{grey}100.4} & {\color{grey}28.4} & {\color{grey}58.3} \\
 & {\color{grey}98.9} & {\color{grey}57.7} & {\color{grey}51.5} & {\color{grey}56.8} & {\color{grey}51.8} & {\color{grey}40.3} & {\color{grey}80.9} & {\color{green2}102.1} & {\color{grey}29.6} & {\color{green2}59.1} \\
 & {\color{grey}97.7} & {\color{grey}57.1} & {\color{grey}50.7} & {\color{grey}56.6} & {\color{grey}51.4} & {\color{grey}39.6} & {\color{grey}78.2} & {\color{grey}101.4} & {\color{grey}29.4} & {\color{grey}58.4} \\
 & {\color{grey}97.9} & {\color{grey}56.8} & {\color{grey}50.3} & {\color{grey}56.0} & {\color{grey}50.8} & {\color{grey}39.7} & {\color{grey}79.1} & {\color{green2}102.8} & {\color{grey}29.6} & {\color{green2}59.4} \\
 & {\color{grey}98.2} & {\color{grey}57.4} & {\color{grey}51.0} & {\color{grey}56.3} & {\color{grey}51.4} & {\color{grey}40.1} & {\color{grey}79.8} & {\color{grey}101.9} & {\color{grey}29.4} & {\color{green2}58.6} \\
\midrule
\multirow{6}{*}{\shortstack{Video-based\\24 views}} & {\color{green2}\underline{101.5}} & {\color{green2}60.2} & {\color{green2}53.3} & {\color{green2}\underline{58.7}} & {\color{green2}\underline{53.4}} & {\color{grey}40.6} & {\color{grey}82.3} & {\color{green2}\textbf{105.0}} & {\color{green2}\textbf{30.3}} & {\color{green2}\textbf{60.0}} \\
 & {\color{green2}100.1} & {\color{green2}59.6} & {\color{green2}52.8} & {\color{green2}58.5} & {\color{green2}53.1} & {\color{grey}40.6} & {\color{grey}80.5} & {\color{green2}102.2} & {\color{grey}29.3} & {\color{green2}58.9} \\
 & {\color{green2}100.7} & {\color{green2}59.9} & {\color{green2}52.9} & {\color{green2}58.4} & {\color{green2}53.0} & {\color{grey}40.9} & {\color{grey}82.9} & {\color{green2}103.1} & {\color{grey}29.4} & {\color{green2}59.2} \\
 & {\color{grey}99.9} & {\color{green2}59.0} & {\color{green2}52.2} & {\color{green2}58.1} & {\color{green2}52.6} & {\color{grey}40.2} & {\color{grey}80.7} & {\color{green2}103.9} & {\color{grey}29.7} & {\color{green2}59.2} \\
 & {\color{grey}100.0} & {\color{green2}58.7} & {\color{green2}52.1} & {\color{green2}57.8} & {\color{green2}52.7} & {\color{grey}40.7} & {\color{grey}83.2} & {\color{green2}102.5} & {\color{grey}29.6} & {\color{green2}58.5} \\
 & {\color{green2}100.0} & {\color{green2}59.3} & {\color{green2}52.8} & {\color{green2}57.5} & {\color{green2}52.4} & {\color{grey}40.0} & {\color{grey}81.0} & {\color{green2}\underline{104.6}} & {\color{green2}\underline{30.2}} & {\color{grey}58.1} \\
\midrule
\multirow{13}{*}{\shortstack{Video-based\\32 views}} & {\color{green2}100.1} & {\color{green2}59.5} & {\color{green2}53.1} & {\color{green2}58.5} & {\color{green2}53.2} & {\color{grey}39.7} & {\color{grey}80.6} & {\color{green2}103.0} & {\color{grey}29.4} & {\color{green2}59.3} \\
 & {\color{green2}\underline{101.2}} & {\color{green2}\textbf{60.9}} & {\color{green2}\textbf{54.1}} & {\color{green2}\underline{59.1}} & {\color{green2}\underline{53.8}} & {\color{grey}40.6} & {\color{grey}82.8} & {\color{green2}102.3} & {\color{grey}29.6} & {\color{green2}58.5} \\
 & {\color{green2}\textbf{101.8}} & {\color{green2}\underline{60.6}} & {\color{green2}\underline{53.9}} & {\color{green2}\underline{59.1}} & {\color{green2}\underline{53.8}} & {\color{green2}\underline{41.5}} & {\color{grey}83.7} & {\color{green2}103.7} & {\color{grey}29.5} & {\color{green2}\underline{59.6}} \\
 & {\color{green2}\underline{101.7}} & {\color{green2}\underline{60.5}} & {\color{green2}53.5} & {\color{green2}\underline{59.2}} & {\color{green2}\underline{53.7}} & {\color{grey}41.0} & {\color{green2}\textbf{85.4}} & {\color{green2}103.2} & {\color{grey}29.8} & {\color{green2}59.2} \\
 & {\color{green2}100.2} & {\color{green2}59.7} & {\color{green2}53.1} & {\color{green2}58.5} & {\color{green2}53.2} & {\color{grey}39.9} & {\color{grey}79.2} & {\color{green2}\underline{104.1}} & {\color{grey}29.9} & {\color{green2}59.1} \\
 & {\color{green2}\underline{101.4}} & {\color{green2}\underline{60.7}} & {\color{green2}\underline{53.9}} & {\color{green2}\textbf{59.3}} & {\color{green2}\textbf{53.8}} & {\color{grey}40.4} & {\color{grey}81.7} & {\color{green2}103.3} & {\color{grey}\underline{30.1}} & {\color{green2}59.2} \\
 & {\color{green2}\underline{101.6}} & {\color{green2}\underline{60.4}} & {\color{green2}\underline{53.7}} & {\color{green2}\underline{59.1}} & {\color{green2}\underline{53.7}} & {\color{grey}40.7} & {\color{grey}83.0} & {\color{green2}\underline{104.3}} & {\color{grey}30.0} & {\color{green2}59.4} \\
 & {\color{green2}\underline{101.5}} & {\color{green2}\underline{60.6}} & {\color{green2}\underline{53.8}} & {\color{green2}\underline{59.2}} & {\color{green2}\underline{53.8}} & {\color{grey}40.7} & {\color{green2}84.1} & {\color{green2}102.9} & {\color{grey}29.6} & {\color{green2}59.2} \\
 & {\color{green2}100.7} & {\color{green2}59.9} & {\color{green2}53.2} & {\color{green2}58.3} & {\color{green2}53.1} & {\color{green2}\underline{41.3}} & {\color{green2}84.0} & {\color{green2}103.5} & {\color{grey}29.3} & {\color{grey}57.8} \\
 & {\color{green2}\underline{101.2}} & {\color{green2}60.1} & {\color{green2}53.4} & {\color{green2}\underline{58.7}} & {\color{green2}\underline{53.4}} & {\color{grey}40.8} & {\color{grey}83.8} & {\color{green2}103.7} & {\color{grey}29.4} & {\color{green2}\underline{59.6}} \\
 & {\color{green2}\underline{101.5}} & {\color{green2}60.0} & {\color{green2}53.3} & {\color{green2}58.6} & {\color{green2}53.1} & {\color{green2}\textbf{41.5}} & {\color{green2}84.4} & {\color{green2}103.6} & {\color{grey}29.8} & {\color{green2}\underline{59.6}} \\
 & {\color{green2}100.3} & {\color{green2}59.4} & {\color{green2}52.8} & {\color{green2}58.3} & {\color{green2}53.0} & {\color{grey}40.6} & {\color{grey}82.5} & {\color{green2}102.6} & {\color{grey}29.4} & {\color{green2}58.9} \\
 & {\color{green2}\underline{100.9}} & {\color{green2}\underline{60.3}} & {\color{green2}53.4} & {\color{green2}57.7} & {\color{green2}52.4} & {\color{grey}41.0} & {\color{grey}83.2} & {\color{green2}103.4} & {\color{grey}29.8} & {\color{green2}\underline{59.6}} \\
\bottomrule

\end{tabular}
}
\end{table}

%
%

\begin{table}[b]
\centering
\caption{Full results for our number of views and subsampled points study (main paper table 5). Only point-based models are shown here. \textbf{Bold} is best. \underline{Underlined} is within 1\% of best.
{\color{green2}Green} is better than SoTA.}
\label{tab:views_points_point_full}
\resizebox{\textwidth}{!}{
\begin{tabular}{l|c|cc|cc|cc|cc|c}
\toprule
Method & All &\multicolumn{2}{c|}{ScanRefer} & \multicolumn{2}{c|}{Multi3DRefer} & \multicolumn{2}{c|}{Scan2Cap} & \multicolumn{2}{c|}{ScanQA} & SQA3D \\
& NS & Ac25 & Ac50 & F$_1$25 & F$_1$50 & B$_4$50 & C50 & C & EM & EM \\

\midrule
\multirow{3}{*}{\shortstack{Point-based (1024 points)\\8 views}} & {\color{grey}88.1} & {\color{grey}48.5} & {\color{grey}43.0} & {\color{grey}49.4} & {\color{grey}45.1} & {\color{grey}37.9} & {\color{grey}70.1} & {\color{grey}93.8} & {\color{grey}27.1} & {\color{grey}56.5} \\
 & {\color{grey}89.9} & {\color{grey}50.4} & {\color{grey}44.5} & {\color{grey}50.0} & {\color{grey}45.4} & {\color{grey}38.2} & {\color{grey}72.3} & {\color{grey}96.5} & {\color{grey}27.5} & {\color{grey}56.9} \\
 & {\color{grey}89.1} & {\color{grey}49.1} & {\color{grey}43.5} & {\color{grey}49.7} & {\color{grey}45.2} & {\color{grey}38.0} & {\color{grey}71.4} & {\color{grey}96.1} & {\color{grey}27.6} & {\color{grey}56.8} \\
\midrule
\multirow{3}{*}{\shortstack{Point-based (1024 points)\\16 views}} & {\color{grey}94.3} & {\color{grey}54.4} & {\color{grey}48.5} & {\color{grey}54.8} & {\color{grey}50.0} & {\color{grey}38.9} & {\color{grey}75.4} & {\color{grey}97.5} & {\color{grey}27.8} & {\color{grey}57.7} \\
 & {\color{grey}95.0} & {\color{grey}54.9} & {\color{grey}48.9} & {\color{grey}54.4} & {\color{grey}49.5} & {\color{grey}39.3} & {\color{grey}77.2} & {\color{grey}98.5} & {\color{grey}28.4} & {\color{grey}57.8} \\
 & {\color{grey}96.4} & {\color{grey}55.8} & {\color{grey}49.9} & {\color{grey}54.9} & {\color{grey}50.1} & {\color{grey}40.3} & {\color{grey}80.6} & {\color{grey}98.1} & {\color{grey}28.5} & {\color{grey}57.8} \\
\midrule
\multirow{4}{*}{\shortstack{Point-based (1024 points)\\24 views}} & {\color{grey}96.6} & {\color{grey}56.2} & {\color{grey}49.7} & {\color{grey}55.9} & {\color{grey}50.7} & {\color{grey}39.3} & {\color{grey}79.3} & {\color{grey}98.8} & {\color{grey}28.9} & {\color{green2}58.6} \\
 & {\color{grey}97.0} & {\color{grey}56.3} & {\color{grey}49.8} & {\color{grey}56.2} & {\color{grey}51.0} & {\color{grey}40.1} & {\color{grey}81.3} & {\color{grey}98.2} & {\color{grey}28.2} & {\color{green2}59.0} \\
 & {\color{grey}96.5} & {\color{grey}56.2} & {\color{grey}49.8} & {\color{grey}56.0} & {\color{grey}50.8} & {\color{grey}39.4} & {\color{grey}78.2} & {\color{grey}99.1} & {\color{grey}28.8} & {\color{grey}57.8} \\
 & {\color{grey}97.0} & {\color{grey}56.5} & {\color{grey}50.2} & {\color{grey}56.3} & {\color{grey}51.2} & {\color{grey}39.7} & {\color{grey}79.0} & {\color{grey}99.1} & {\color{grey}28.7} & {\color{grey}58.3} \\
\midrule
\multirow{3}{*}{\shortstack{Point-based (1024 points)\\32 views}} & {\color{grey}96.5} & {\color{grey}56.8} & {\color{grey}50.2} & {\color{grey}56.4} & {\color{grey}51.2} & {\color{grey}39.0} & {\color{grey}77.2} & {\color{grey}98.8} & {\color{grey}28.4} & {\color{green2}58.5} \\
 & {\color{grey}96.6} & {\color{grey}56.8} & {\color{grey}50.3} & {\color{grey}56.3} & {\color{grey}50.9} & {\color{grey}39.0} & {\color{grey}77.5} & {\color{grey}98.6} & {\color{grey}28.7} & {\color{green2}58.6} \\
 & {\color{grey}97.2} & {\color{grey}56.9} & {\color{grey}50.4} & {\color{grey}57.1} & {\color{grey}51.9} & {\color{grey}39.6} & {\color{grey}78.6} & {\color{grey}99.5} & {\color{grey}28.6} & {\color{grey}57.7} \\
\midrule
\multirow{3}{*}{\shortstack{Point-based (2048 points)\\8 views}} & {\color{grey}90.1} & {\color{grey}49.7} & {\color{grey}44.1} & {\color{grey}50.0} & {\color{grey}45.6} & {\color{grey}38.2} & {\color{grey}74.4} & {\color{grey}95.7} & {\color{grey}27.8} & {\color{grey}57.3} \\
 & {\color{grey}90.9} & {\color{grey}51.3} & {\color{grey}45.4} & {\color{grey}50.7} & {\color{grey}46.0} & {\color{grey}37.9} & {\color{grey}74.8} & {\color{grey}96.5} & {\color{grey}28.0} & {\color{grey}57.0} \\
 & {\color{grey}90.2} & {\color{grey}50.0} & {\color{grey}44.1} & {\color{grey}49.7} & {\color{grey}45.0} & {\color{grey}38.8} & {\color{grey}75.1} & {\color{grey}96.7} & {\color{grey}28.0} & {\color{grey}56.7} \\
\midrule
\multirow{3}{*}{\shortstack{Point-based (2048 points)\\16 views}} & {\color{grey}96.2} & {\color{grey}55.4} & {\color{grey}49.4} & {\color{grey}55.4} & {\color{grey}50.4} & {\color{grey}39.6} & {\color{grey}77.4} & {\color{grey}100.2} & {\color{grey}29.0} & {\color{grey}58.2} \\
 & {\color{grey}96.6} & {\color{grey}55.2} & {\color{grey}49.3} & {\color{grey}55.4} & {\color{grey}50.6} & {\color{grey}40.4} & {\color{grey}80.6} & {\color{grey}98.4} & {\color{grey}28.7} & {\color{green2}58.7} \\
 & {\color{grey}97.0} & {\color{grey}55.6} & {\color{grey}49.5} & {\color{grey}55.1} & {\color{grey}50.1} & {\color{grey}39.7} & {\color{grey}79.6} & {\color{green2}\underline{102.5}} & {\color{grey}29.7} & {\color{grey}58.2} \\
\midrule
\multirow{3}{*}{\shortstack{Point-based (2048 points)\\24 views}} & {\color{grey}98.5} & {\color{grey}57.8} & {\color{grey}51.1} & {\color{grey}57.0} & {\color{grey}51.9} & {\color{grey}39.7} & {\color{grey}81.2} & {\color{grey}100.8} & {\color{grey}28.9} & {\color{green2}59.7} \\
 & {\color{grey}98.9} & {\color{grey}57.7} & {\color{grey}51.1} & {\color{grey}57.1} & {\color{grey}51.9} & {\color{grey}40.4} & {\color{grey}81.6} & {\color{green2}\textbf{103.0}} & {\color{grey}29.4} & {\color{grey}58.3} \\
 & {\color{grey}98.4} & {\color{grey}57.2} & {\color{grey}50.5} & {\color{grey}56.7} & {\color{grey}51.4} & {\color{grey}40.1} & {\color{grey}81.5} & {\color{grey}101.4} & {\color{grey}29.2} & {\color{green2}59.6} \\
\midrule
\multirow{3}{*}{\shortstack{Point-based (2048 points)\\32 views}} & {\color{grey}98.6} & {\color{grey}58.0} & {\color{grey}51.5} & {\color{green2}57.5} & {\color{grey}52.0} & {\color{grey}39.5} & {\color{grey}80.0} & {\color{grey}100.4} & {\color{grey}29.1} & {\color{green2}59.8} \\
 & {\color{grey}98.5} & {\color{grey}57.8} & {\color{grey}51.2} & {\color{grey}57.2} & {\color{grey}51.9} & {\color{grey}39.4} & {\color{grey}81.3} & {\color{grey}99.8} & {\color{grey}29.2} & {\color{green2}59.5} \\
 & {\color{grey}98.8} & {\color{grey}58.0} & {\color{grey}51.3} & {\color{green2}57.7} & {\color{green2}52.2} & {\color{grey}39.6} & {\color{grey}81.2} & {\color{grey}101.4} & {\color{grey}29.1} & {\color{green2}59.0} \\
\midrule
\multirow{3}{*}{\shortstack{Point-based (4096 points)\\8 views}} & {\color{grey}90.9} & {\color{grey}51.0} & {\color{grey}45.2} & {\color{grey}50.6} & {\color{grey}46.2} & {\color{grey}38.9} & {\color{grey}74.7} & {\color{grey}95.9} & {\color{grey}27.4} & {\color{grey}57.2} \\
 & {\color{grey}91.1} & {\color{grey}51.3} & {\color{grey}45.3} & {\color{grey}50.5} & {\color{grey}45.8} & {\color{grey}38.9} & {\color{grey}74.5} & {\color{grey}97.5} & {\color{grey}28.2} & {\color{grey}56.3} \\
 & {\color{grey}91.5} & {\color{grey}51.1} & {\color{grey}44.9} & {\color{grey}50.8} & {\color{grey}46.1} & {\color{grey}38.8} & {\color{grey}74.9} & {\color{grey}97.4} & {\color{grey}28.5} & {\color{grey}57.5} \\
\midrule
\multirow{3}{*}{\shortstack{Point-based (4096 points)\\16 views}} & {\color{grey}96.6} & {\color{grey}55.9} & {\color{grey}49.7} & {\color{grey}55.7} & {\color{grey}50.6} & {\color{grey}39.1} & {\color{grey}79.0} & {\color{grey}99.9} & {\color{grey}28.6} & {\color{green2}59.3} \\
 & {\color{grey}98.0} & {\color{grey}56.2} & {\color{grey}50.0} & {\color{grey}55.9} & {\color{grey}50.9} & {\color{grey}40.4} & {\color{grey}81.4} & {\color{grey}101.9} & {\color{grey}29.6} & {\color{green2}59.0} \\
 & {\color{grey}97.6} & {\color{grey}56.7} & {\color{grey}50.5} & {\color{grey}55.9} & {\color{grey}50.7} & {\color{grey}40.3} & {\color{grey}81.2} & {\color{grey}100.2} & {\color{grey}29.0} & {\color{green2}58.7} \\
\midrule
\multirow{3}{*}{\shortstack{Point-based (4096 points)\\24 views}} & {\color{grey}99.3} & {\color{grey}57.8} & {\color{grey}51.4} & {\color{green2}57.5} & {\color{grey}52.1} & {\color{grey}40.5} & {\color{green2}84.3} & {\color{grey}100.4} & {\color{grey}29.1} & {\color{green2}59.4} \\
 & {\color{green2}100.1} & {\color{grey}58.0} & {\color{grey}51.4} & {\color{green2}57.5} & {\color{grey}52.1} & {\color{green2}\underline{41.4}} & {\color{green2}\underline{86.3}} & {\color{grey}101.8} & {\color{grey}29.5} & {\color{green2}59.3} \\
 & {\color{green2}100.1} & {\color{green2}58.4} & {\color{grey}51.7} & {\color{green2}57.9} & {\color{green2}52.5} & {\color{grey}40.8} & {\color{grey}83.6} & {\color{green2}\underline{102.9}} & {\color{grey}29.8} & {\color{green2}58.8} \\
\midrule
\multirow{8}{*}{\shortstack{Point-based (4096 points)\\32 views}} & {\color{green2}\underline{100.5}} & {\color{green2}59.0} & {\color{green2}52.4} & {\color{green2}57.8} & {\color{green2}52.4} & {\color{green2}\textbf{41.4}} & {\color{green2}84.9} & {\color{green2}\underline{102.6}} & {\color{grey}29.3} & {\color{green2}59.2} \\
 & {\color{green2}100.2} & {\color{green2}59.0} & {\color{green2}52.2} & {\color{green2}58.2} & {\color{green2}\underline{52.8}} & {\color{grey}40.5} & {\color{green2}84.0} & {\color{grey}101.4} & {\color{grey}29.2} & {\color{green2}\underline{60.0}} \\
 & {\color{green2}\underline{100.9}} & {\color{green2}\textbf{60.2}} & {\color{green2}\textbf{53.5}} & {\color{green2}\underline{58.3}} & {\color{green2}\underline{52.8}} & {\color{grey}40.6} & {\color{green2}84.3} & {\color{green2}\underline{102.8}} & {\color{grey}29.5} & {\color{green2}59.4} \\
 & {\color{grey}100.0} & {\color{green2}58.9} & {\color{green2}52.0} & {\color{green2}57.9} & {\color{green2}52.3} & {\color{grey}40.4} & {\color{green2}84.0} & {\color{grey}101.2} & {\color{grey}29.4} & {\color{green2}59.8} \\
 & {\color{green2}\underline{101.0}} & {\color{green2}58.9} & {\color{green2}52.3} & {\color{green2}\underline{58.5}} & {\color{green2}\underline{53.1}} & {\color{grey}\underline{41.1}} & {\color{green2}\textbf{86.9}} & {\color{grey}101.5} & {\color{grey}29.4} & {\color{green2}\underline{60.1}} \\
 & {\color{green2}\underline{100.6}} & {\color{green2}59.3} & {\color{green2}52.6} & {\color{green2}\underline{58.5}} & {\color{green2}\underline{53.1}} & {\color{grey}40.8} & {\color{green2}85.1} & {\color{grey}101.8} & {\color{grey}29.0} & {\color{green2}59.7} \\
 & {\color{green2}\underline{100.5}} & {\color{green2}59.1} & {\color{green2}52.4} & {\color{green2}57.8} & {\color{green2}52.5} & {\color{grey}40.7} & {\color{green2}\underline{86.0}} & {\color{grey}101.1} & {\color{grey}29.3} & {\color{green2}\underline{59.9}} \\
 & {\color{green2}\underline{100.6}} & {\color{green2}59.1} & {\color{green2}52.5} & {\color{green2}\underline{58.2}} & {\color{green2}\underline{52.9}} & {\color{grey}40.5} & {\color{green2}84.3} & {\color{grey}101.8} & {\color{grey}29.4} & {\color{green2}\textbf{60.5}} \\
\midrule
\multirow{3}{*}{\shortstack{Point-based (8192 points)\\8 views}} & {\color{grey}90.6} & {\color{grey}50.7} & {\color{grey}45.0} & {\color{grey}50.8} & {\color{grey}46.3} & {\color{grey}39.0} & {\color{grey}74.8} & {\color{grey}95.2} & {\color{grey}27.3} & {\color{grey}56.4} \\
 & {\color{grey}91.7} & {\color{grey}51.1} & {\color{grey}45.2} & {\color{grey}50.7} & {\color{grey}46.1} & {\color{grey}39.2} & {\color{grey}75.4} & {\color{grey}98.6} & {\color{grey}28.5} & {\color{grey}57.0} \\
 & {\color{grey}91.1} & {\color{grey}50.3} & {\color{grey}44.3} & {\color{grey}50.0} & {\color{grey}45.3} & {\color{grey}39.5} & {\color{grey}75.9} & {\color{grey}97.1} & {\color{grey}28.2} & {\color{grey}57.8} \\
\midrule
\multirow{3}{*}{\shortstack{Point-based (8192 points)\\16 views}} & {\color{grey}97.7} & {\color{grey}56.5} & {\color{grey}50.5} & {\color{grey}55.9} & {\color{grey}51.0} & {\color{grey}40.1} & {\color{grey}80.6} & {\color{grey}101.2} & {\color{grey}29.3} & {\color{grey}58.1} \\
 & {\color{grey}98.6} & {\color{grey}56.6} & {\color{grey}50.2} & {\color{grey}56.2} & {\color{grey}51.0} & {\color{grey}40.7} & {\color{grey}83.3} & {\color{green2}\underline{102.9}} & {\color{grey}29.6} & {\color{green2}58.5} \\
 & {\color{grey}97.3} & {\color{grey}56.0} & {\color{grey}49.7} & {\color{grey}55.4} & {\color{grey}50.1} & {\color{grey}40.0} & {\color{grey}81.9} & {\color{grey}100.6} & {\color{grey}29.0} & {\color{green2}59.4} \\
\midrule
\multirow{3}{*}{\shortstack{Point-based (8192 points)\\24 views}} & {\color{green2}100.4} & {\color{green2}58.7} & {\color{green2}52.1} & {\color{green2}57.9} & {\color{green2}52.5} & {\color{grey}40.7} & {\color{green2}84.9} & {\color{green2}\underline{102.2}} & {\color{grey}29.7} & {\color{green2}59.6} \\
 & {\color{green2}100.2} & {\color{green2}58.5} & {\color{green2}52.0} & {\color{green2}57.6} & {\color{green2}52.5} & {\color{grey}40.5} & {\color{green2}84.8} & {\color{green2}\underline{102.1}} & {\color{grey}29.5} & {\color{green2}59.8} \\
 & {\color{grey}99.6} & {\color{grey}57.9} & {\color{grey}51.3} & {\color{green2}57.6} & {\color{green2}52.3} & {\color{grey}40.3} & {\color{grey}83.8} & {\color{grey}101.8} & {\color{grey}29.4} & {\color{green2}\underline{60.1}} \\
\midrule
\multirow{4}{*}{\shortstack{Point-based (8192 points)\\32 views}} & {\color{green2}\textbf{101.5}} & {\color{green2}\underline{59.7}} & {\color{green2}52.8} & {\color{green2}\textbf{58.8}} & {\color{green2}\textbf{53.2}} & {\color{grey}40.7} & {\color{green2}\underline{86.8}} & {\color{green2}\underline{102.1}} & {\color{grey}29.8} & {\color{green2}\underline{60.3}} \\
 & {\color{green2}\underline{101.1}} & {\color{green2}\underline{59.7}} & {\color{green2}\underline{53.1}} & {\color{green2}\underline{58.4}} & {\color{green2}\underline{53.0}} & {\color{grey}40.9} & {\color{green2}85.5} & {\color{green2}\underline{102.1}} & {\color{grey}\underline{29.9}} & {\color{green2}59.3} \\
 & {\color{green2}100.4} & {\color{green2}58.4} & {\color{grey}51.7} & {\color{green2}57.8} & {\color{green2}52.5} & {\color{grey}40.9} & {\color{green2}\underline{86.2}} & {\color{green2}\underline{102.4}} & {\color{grey}29.8} & {\color{green2}59.1} \\
 & {\color{green2}\underline{101.0}} & {\color{green2}59.2} & {\color{green2}52.4} & {\color{green2}58.1} & {\color{green2}\underline{52.7}} & {\color{grey}\underline{41.2}} & {\color{green2}\underline{86.2}} & {\color{green2}\underline{103.0}} & {\color{grey}29.6} & {\color{green2}59.3} \\
\bottomrule

\end{tabular}
}
\end{table}

\end{document}